%% file: main.tex
\documentclass[3p,twocolumn]{elsarticle}
\usepackage{times}
\usepackage{url}
\usepackage{amsmath}
\usepackage{graphicx}
\usepackage{array}
\usepackage{booktabs}
\usepackage{multirow}
\usepackage{tcolorbox}
\usepackage{lineno,hyperref}
\modulolinenumbers[5]
\newcommand{\minus}{\scalebox{0.75}[1.0]{$-$}}
\usepackage{subcaption}
\newpageafter{abstract}

\begin{document}

\begin{frontmatter}

\title{Time-Aware Evidence Ranking for Fact-Checking}

\author[mymainaddress,mysecondaryaddress]{Liesbeth Allein\corref{mycorrespondingauthor}}
\cortext[mycorrespondingauthor]{Corresponding author}
\ead{Liesbeth.ALLEIN@ec.europa.eu}

\author[thirdaddress]{Isabelle Augenstein}

\author[mysecondaryaddress]{Marie-Francine Moens}

\address[mymainaddress]{European Commission, Joint Research Centre (JRC), Ispra, Italy}
\address[mysecondaryaddress]{Department of Computer Science, KU Leuven, Celestijnenlaan 200A, 3001 Leuven, Belgium}
\address[thirdaddress]{Department of Computer Science, University of Copenhagen, Universitetsparken 1, 2100 Copenhagen, Denmark}





\begin{abstract}
Truth can vary over time. Fact-checking decisions on claim veracity should therefore take into account temporal information of both the claim and supporting or refuting evidence. In this work, we investigate the hypothesis that the timestamp of a Web page is crucial to how it should be ranked for a given claim. We delineate four temporal ranking methods that constrain evidence ranking differently and simulate hypothesis-specific evidence rankings given the evidence timestamps as gold standard. Evidence ranking in three fact-checking models is ultimately optimized using a learning-to-rank loss function. Our study reveals that time-aware evidence ranking not only surpasses relevance assumptions based purely on semantic similarity or position in a search results list, but also improves veracity predictions of time-sensitive claims in particular. 
\end{abstract}


\end{frontmatter}


\input{introduction.tex}
\input{related_work}
\input{time}
\input{experiments}

\input{results}
\input{discussion}
\input{conclusion}

\section*{Acknowledgements}
This work was realised with the collaboration of the European Commission Joint Research Centre under the Collaborative Doctoral Partnership Agreement No 35332 and has been supported by COST Action CA18231. The research was designed, the base model was implemented and a first manuscript was written when Liesbeth Allein was at KU Leuven; the additional models were implemented, further analyses were performed and the manuscript was finalized when she was at the European Commission. The scientific output expressed does not imply a policy position of the European Commission. Neither the European Commission nor any person acting on behalf of the Commission is responsible for the use which might be made of this publication.

\bibliography{bibliography}

\section*{Appendix}
\input{supplementary_material}
\input{supplementary_material_B}

\end{document}

%% file: introduction.tex
\section{Introduction}

\begin{figure}
    \centering
    \includegraphics[width=8cm]{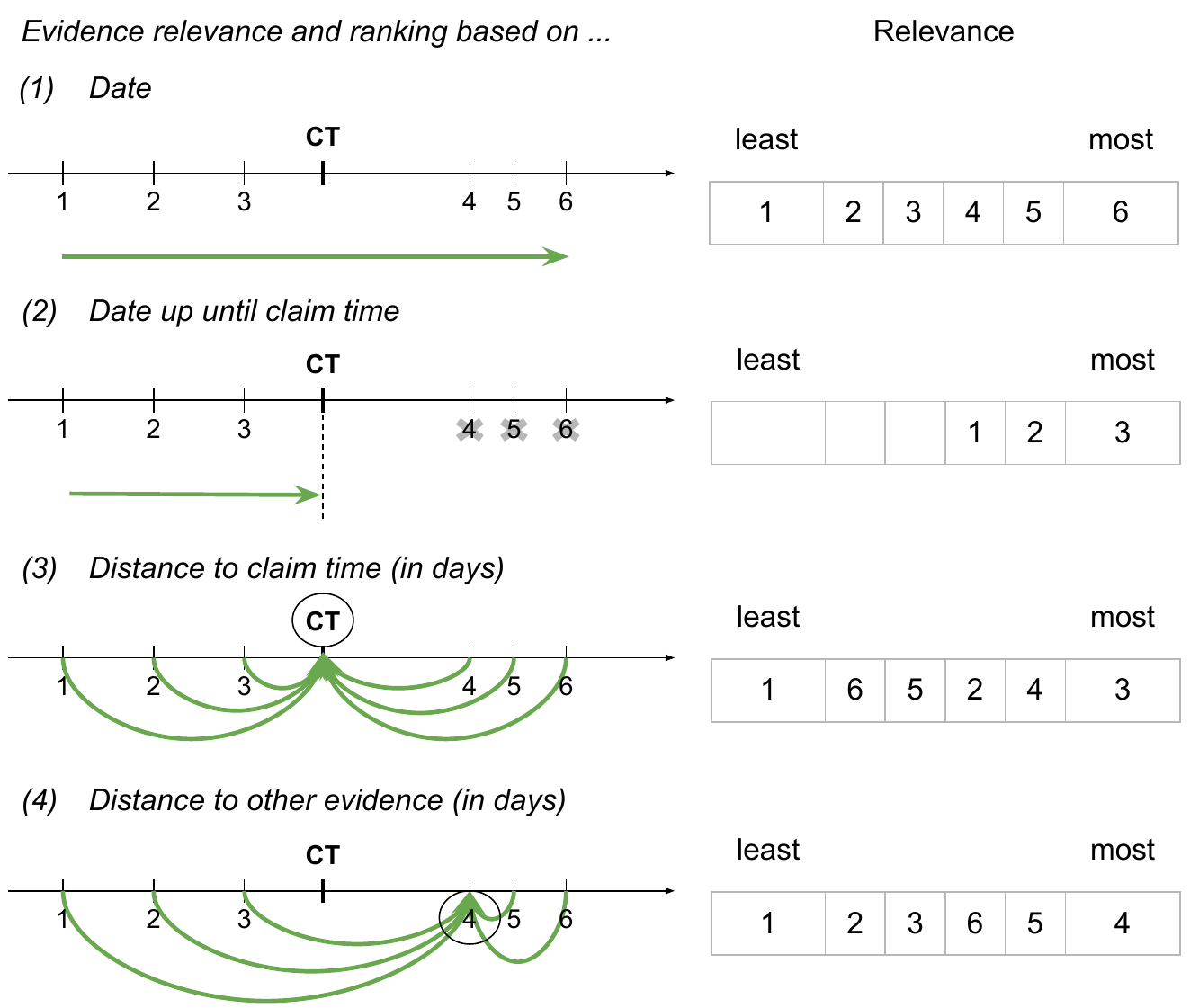}
    \caption{Given a claim and the date at which the claim was published (claim time; CT), the relevance estimation and ranking of evidence snippets positioned on the same timeline as the claim (1-6) can follow various assumptions.} 
    \label{fig:introduction_image}
\end{figure}

While some claims are incontestably true or false at any time (e.g. \textit{``Smoking increases the risk of cancer''}), the veracity of others is subject to time indications and temporal dynamics (e.g. \textit{``Face masks are obligatory on public transport''}) \citep{halpin2008temporal}. Not only their veracity, but also their semantics are time-sensitive or time-dependent as connotations and real-world references can change over time. Evidence supporting or refuting such time-sensitive claims is likewise time-dependent. The relevance of evidence documents, which reflects both their semantic relatedness to the claim and suitability for accurate claim veracity prediction, is thus relative to a claim's publication date and/or the documents' publication date. A fact-checking model's inability to correctly frame both claim and evidence in time and its inability to rank evidence documents by relevance can result in inaccurate semantic representations, truth predictions and relevance estimations. Nonetheless, automated fact-checking research has paid little attention to the temporal dynamics of truth, semantics and relevance. In this work, we focus on the temporal relevance of Web documents that serve as evidence to a given claim. We introduce four temporal ranking methods which constrain evidence ranking relying on diverse hypotheses for evidence relevance and explore how time-aware evidence ranking impacts the veracity prediction performance of three fact-checking models (Figure \ref{fig:introduction_image}).

The first two methods simply rank evidence by date in descending order: \textit{evidence-based date ranking} sorts all given evidence, while \textit{claim-based date ranking} only ranks evidence published before and at claim time. The other two methods rank evidence by distance in days to either the claim (\textit{claim-centered distance ranking}) or the other evidence in the same set (\textit{evidence-centered distance ranking}), both in ascending order. 
These methods then simulate method-specific ground-truth rankings given the timestamp of each evidence snippet.
Ultimately, the evidence ranking module of a fact-checking model is directly optimized using a dedicated learning-to-rank loss function, which measures the agreement between the model's ranking output and the simulated ground-truth rankings. 

In summary, the \textbf{contributions} of this work are as follows.
\begin{itemize}
    \item We propose to model the temporal dynamics of evidence for content-based fact-checking and show that it outperforms a ranking of Web documents based purely on semantic similarity -- as used in prior work -- and search engine ranking.
    \item We test various hypotheses for evidence relevance using timestamps and explore the performance differences between those hypotheses.
    \item We train evidence ranking by optimizing a learning-to-rank loss function. This elegant, yet effective approach requires only a few adjustments to the model architecture and can be easily added to any fact-checking model.
    \item Optimizing evidence ranking using a dedicated learning-to-rank loss function is, to our knowledge, novel in automated fact-checking research.
\end{itemize}

%% file: related_work.tex
\section{Related Work}

Previous work on content-based fact-checking that exploits both claim and evidence has differentiated between pieces of evidence in various manners. Some consider different evidence documents to be equally important \citep{mishra2019sadhan,thorne2018fever}. Others weigh or rank evidence according to its assumed relevance to the claim. \citet{liu2019kernel}, for instance, link evidence relevance to node and edge kernel importance in an evidence graph using neural matching kernels. \citet{li2016verification,li2011t} define evidence relevance in terms of evidence position in a search engine's ranking list, while \citet{wang2015multi} relate it with source popularity. Another line of work views evidence documents as interconnected chains and selects them based on their semantic connections, investigating multi-hop fact checking \cite{conf/iclr/ZhaoXRSBT20,Ostrowski2021ijcai}. However, evidence relevance has been principally associated with semantic similarity between claim-evidence pairs and is computed using language models \citep{zhong2019reasoning}, textual entailment/inference models \citep{hanselowski2018ukp,zhou2019gear,nie2019combining}, cosine similarity \citep{miranda2019automated} or token matching and sentence position in the evidence document \citep{yoneda2018ucl}. In contrast to previous work, we hypothesize that the timestamp of a piece of evidence and reasoning with the temporal information are crucial to how evidence relevance should be defined and how it should be ranked for a given claim. 
\\ \\
The dynamics of time in fact-checking have not been widely studied yet. \citet{yamamoto2008supporting} incorporate the idea of temporal factuality in a fact-checking model. Uncertain facts are input as queries to a search engine, and a fact's trustworthiness is determined based on the detected sentiment and frequency of alternative and counter facts in the search results in a given time frame. 
However, the authors point out that the frequency of a fact can be misleading, with incorrect claims possibly having more hits than correct ones \citep{li2011t}. \citet{hidey2020deseption} recently published an adversarial dataset that can be used to evaluate a fact-checking model's temporal reasoning abilities. In this dataset, arithmetic, range and verbalized time indications are altered using date manipulation heuristics.
\citet{zhou-etal-2020-probabilistic} study pattern-based temporal fact extraction. By first extracting temporal facts from a corpus of unstructured texts using textual pattern-based methods, they model pattern reliability based on time cues, such as text generation timestamps and in-text temporal tags. Unreliable and incorrect temporal facts are then automatically discarded. However, relying on the above method, a large amount of data is needed to determine a claim's veracity, which might not be available for new claims yet.

%% file: time.tex
\newcommand{\abs}[1]{\lvert #1 \rvert}
\newtheorem{theorem}{Hypothesis}

\section{Time-Aware Evidence Ranking}\label{ranking_methods}

We encourage a content-based fact-checking model to reason about the temporal semantics and time dependency of both claim and evidence by constraining the model's ranking module. This module assigns a relevance/ranking score to each Web document serving as evidence. During training, the ranking module is optimized using a learning-to-rank loss function, which measures the agreement between the learned ranking output and the expected ranking. As ground-truth evidence rankings are lacking, we introduce four ranking methods relying on several hypotheses on temporal relevance and use these methods to simulate ground-truth rankings. 

We first discuss the three fact-checking models whose ranking module we will optimize on the simulated ground-truth evidence rankings. Next, we elaborate on the specific learning-to-rank loss used during training. We then explain how timestamps for a given claim and evidence set are extracted and normalized. Finally, we introduce the four temporal ranking methods that (a) constrain evidence ranking following several hypotheses for evidence relevance and (b) simulate hypothesis-specific ground-truth evidence rankings. From Section 3.3 onwards, we illustrate the time extraction/normalization process and the temporal ranking methods with a sample evidence set and claim taken from the MultiFC dataset \cite{augenstein2019multifc}.

\input{model_architecture}

\subsection{Learning-to-Rank Loss}

In order to optimize evidence ranking, we need a loss function that measures how correctly an evidence snippet is ranked with regard to the other snippets in the same evidence set. For this, the \textbf{ListMLE} loss \citep{xia2008listwise} is computed:
\begin{equation}\label{ListMLE}
     ListMLE(r(E),R) = -\sum _{i=1}^{N}log P(R_i|E_i;r)
\end{equation}
\begin{equation}\label{probability}
   P(R_i|E_i;r) = \prod_{u=1}^{K} \frac{exp(r(E_{R_{i_u}}))}{\sum_{v=u}^{K} exp(r(E_{R_{i_v}}))}
\end{equation}
ListMLE is a listwise, non-measure-specific learning-to-rank algorithm that uses the negative log-likelihood of a ground-truth permutation as loss function \citep{liu2011learning}. It is based on the Plackett-Luce model, that is, the probability of a permutation is first decomposed into the product of a stepwise conditional probability, with the $u$-th conditional probability standing for the probability that the snippet is ranked at the $u$-th position given that the top $u$ \text{-} 1 snippets are ranked correctly. ListMLE is used for optimizing the evidence ranking with each of the four temporal ranking methods.
\\ \\
In case an evidence snippet is excluded from the ground-truth evidence ranking \(R_i\) or lacks a timestamp,
we apply a mask over the predicted evidence ranking vector and compute the ListMLE loss over the ranking scores of the included evidence snippets with timestamps. We assume that the direct optimization of these evidence snippets' ranking scores will indirectly influence the ranking scores of the others as they may contain similar explicit time references further in their text or exhibit similar patterns.

We are, to our knowledge, the first to specifically constrain evidence ranking in automated fact-checking using a dedicated, learning-to-rank loss on simulated ground-truth time-dependent rankings.

\subsection{Temporal Relevance and Ranking Methods}

\begin{figure*}
    \centering
    \includegraphics[width=12cm]{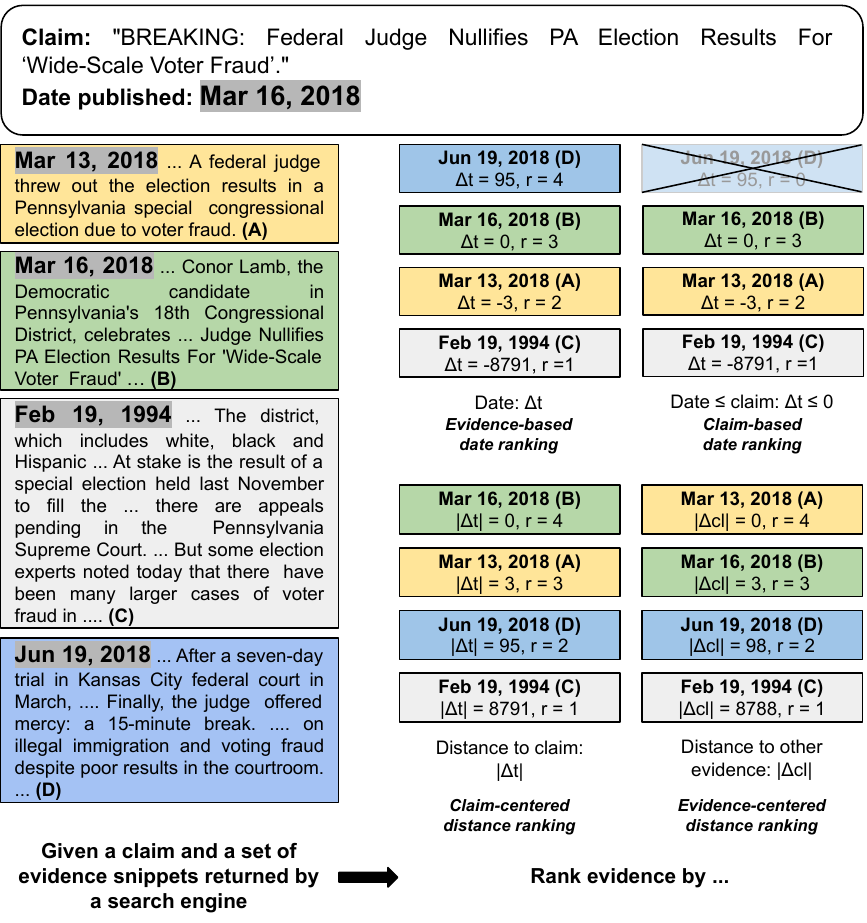}
    \caption{Overview of our temporal ranking methods that constrain evidence ranking following different hypotheses for temporal relevance: \textit{evidence-based date ranking}, \textit{claim-based date ranking}, \textit{claim-centered distance ranking} and \textit{evidence-centered distance ranking}. The example is taken from the MultiFC dataset \cite{augenstein2019multifc}. Although the methods actually rank up to ten snippets per claim in the experiments, we display four snippets for simplicity. The evidence timestamps are first extracted and normalized to their distance to the claim in days (${\Delta}t$). In the fourth ranking method, \textit{evidence-centered distance ranking}, the mediod of the evidence set and the snippets' distance to that mediod is computed (${\Delta}cl$). Based on the ${\Delta}t$ or ${\Delta}cl$ values, the temporal ranking methods assign a ranking score $r$ to each evidence snippet. A higher score denotes a higher degree of relevance.} 
    \label{fig:graphical_image}
\end{figure*}

We explain how temporal information for both claim and evidence is extracted and normalized. We then introduce the four temporal ranking methods that rank the Web documents in the evidence set. For illustration purposes, we use a claim and evidence set from the MultiFC dataset \cite{augenstein2019multifc}:

\(E_s\) = \{
\begin{itemize}
    \item[$e_{s_1}$]  \textit{``Mar  13,  2018 ... A federal judge threw  out  the  election  results  in  a Pennsylvania special  congressional election due to voter fraud."} 
    \item[$e_{s_2}$]  \textit{``Mar 16, 2018... Conor Lamb, the Democratic candidate in Pennsylvania's  18th  Congressional District, celebrates ... Judge Nullifies PA Election Results For 'Wide-Scale Voter  Fraud' ..."} 
    \item[$e_{s_3}$]  \textit{``Feb  19,  1994...  The  district, which  includes  white,  black  and Hispanic ... At stake is the result of a special  election  held  last  November to  fill  the  ...  there  are  appeals pending   in   the      Pennsylvania Supreme Court. ... But some election experts noted today that there  have been  many  larger  cases  of  voter fraud in ...."}
    \item[$e_{s_4}$]  \textit{``Jun 19, 2018... After a seven-day trial  in  Kansas  City  federal  court  in March, .... Finally, the judge  offered mercy:  a  15-minute  break.  ....  on illegal  immigration  and  voting  fraud despite poor results in the courtroom. ..."}
\end{itemize}

\} and \(c_s\) = \textit{``BREAKING:  Federal  Judge  Nullifies  PA  Election  Results  For ‘Wide-Scale Voter Fraud’."} with timestamp \textit{``Mar 16, 2018"} given as claim metadata. A graphical overview of the temporal ranking methods applied to the sample claim and evidence set is provided in Figure \ref{fig:graphical_image}.

\subsection{Timestamp Extraction and Normalization}

The dataset contains \(N\) claim sentences \(c_i \in C\) and \(N\) evidence sets \(E_i \in E\), each providing \(K \leq 10\) evidence snippets \(e_{i_j} \in E_i\). A timestamp for each claim \(c_i\) is included as metadata. For evidence snippet \(e_{i_{j}}\), however, we need to extract the timestamp from the evidence text itself. A publication timestamp is frequently given at the beginning of the text, which facilitates timestamp extraction. We then normalize all extracted timestamps to \textit{year-month-day}, resulting in:
\begin{equation}
    t:C,E \xrightarrow{} C^t = \{t(c_1),...,t(c_N)\}, E^t = \{E^t_i,...,E^t_N\}
\end{equation}
with \(E^t_i =  \{t(e_{i_{1}}),...,t(e_{i_{K}})\}\). For each evidence snippet \(e_{i_{j}} \in E_i\), the temporal distance in days between claim time \(t(c_i)\) and evidence time \(t(e_{i_j})\) is calculated by subtracting \(t(c_i)\) from \(t(e_{i_j})\). This results in:
\begin{equation}
    {\Delta}t: E^t \xrightarrow{} {\Delta}E^t = \{{\Delta}E^t_i,...,{\Delta}E^t_N\} 
\end{equation}
with \({\Delta}E^t_i =  \{{\Delta}t(e_{i_{1}}),...,{\Delta}t(e_{i_{K}})\}\). Positive and negative \({\Delta}t(e_{i_{j}})\) denote evidence snippets that are published later and earlier than \(t(c_i)\), respectively. 
These integer time values are subsequently used for simulating the ground-truth evidence rankings.
\\ \\
\textbf{Example:}
For \(E_s = \{e_{s_1}, e_{s_2}, e_{s_3}, e_{s_4}\}\) and \(c_s\):
\begin{itemize}
    \item Extract timestamps from evidence texts: \\ \(E_s = \{\textit{``Mar  13,  2018"}, \textit{``Mar 16, 2018"}, \\ \textit{``Feb  19,  1994"}, \textit{``Jun 19, 2018"}\}\)
    \item Normalize all timestamps to \textit{year-month-day} (4): \(t: c_s, E_s \xrightarrow{} t(c_s) = \textit{2018{\minus}3{\minus}16}, \\ E^t_s = \{\textit{2016{\minus}3{\minus}13}, \textit{2018{\minus}3{\minus}16}, \textit{1994{\minus}2{\minus}19}, \\ \textit{2018{\minus}6{\minus}19}\}\) 
    \item Calculate the temporal distance in days to \(t(c_s)\) for each \(t(e_{s_j}) \in E^t_s\) (5): \\ \({\Delta}t: E^t_s \xrightarrow{} {\Delta}E^t_s = \{{\minus}3, 0, {\minus}8791, 95\}\)
\end{itemize}

\subsection{Temporal Ranking Methods}

\begin{table*}[!htb]
    \fontsize{9}{8}\selectfont
    \centering
    \begin{tabular}{ccc}
        \toprule
       \textbf{Temporal} & \textbf{Ranking Constraint} & \textbf{Ranking} \\
       \textbf{Ranking Method} & \(\forall e_{i_{j}},e_{i_{k}} \in E_i:\) & \textbf{ \(R_s\)} \\
        \toprule
        \textit{Evidence-based} & \multirow{2}{*}{\({\Delta}t(e_{i_{j}}) \leq {\Delta}t(e_{i_{k}}) \implies r(e_{i_{j}}) \leq r(e_{i_{k}})\)} & \multirow{2}{*}{\(\{2,3,1,4\}\)} \\
        date ranking & & \\
        \textit{Claim-based}  & \(({\Delta}t(e_{i_{j}}) \leq {\Delta}t(e_{i_{k}})) \land ({\Delta}t(e_{i_{j}}) \land {\Delta}t(e_{i_{k}}) \leq 0)\) & \multirow{2}{*}{\(\{2,3,1,0\}\)} \\
        date ranking & \(\implies r(e_{i_{j}}) \leq r(e_{i_{k}})\) & \\
        \midrule
        \textit{Claim-centered} & \multirow{2}{*}{\(\abs{{\Delta}t(e_{i_{j}})} \geq \abs{{\Delta}t(e_{i_{k}})} \implies r(e_{i_{j}}) \leq r(e_{i_{k}})\)} & \multirow{2}{*}{\(\{3,4,1,2\}\)} \\
        distance ranking & & \\
        \textit{Evidence-centered} & \multirow{2}{*}{\(\abs{{\Delta}cl(e_{i_{j}})} \geq \abs{{\Delta}cl(e_{i_{k}})} \implies r(e_{i_{j}}) \leq r(e_{i_{k}})\)} & \multirow{2}{*}{\(\{4,3,1,2\}\)} \\
        distance ranking & & \\
        \bottomrule
    \end{tabular}
    \caption{Overview of the four temporal ranking methods and their ranking constraints. The last column provides sample ground-truth rankings for a given evidence set \(E_s = \{e_{s_1},e_{s_2},e_{s_3},e_{s_4}\}\) with \({\Delta}E^t_s = \{\minus3,0,\minus8791,95\}\) and \({\Delta}E_s^{cl} = \{0,3,8788,98\}\). A higher ranking score denotes higher relevance.} 
    \label{tab:methods_overview}
\end{table*}

All four temporal ranking methods assign ranking scores to the evidence snippets following a method-specific ranking constraint. 
\begin{equation}
    r:E \xrightarrow{} R = \{R_i,...,R_N\}
\end{equation}
with \(R_i =  \{r(e_{i_{1}}),...,r(e_{i_{K}})\}\). In all methods, an evidence snippet's ranking score depends on its own position in time and that of the other evidence snippets in the same evidence set. The ranking scores are proportional to relevance, i.e., a higher ranking score denotes higher relevance. The ranking methods fall into two categories: date-based and distance-based ranking \citep{kanhabua2016temporal,bansal2019temporal}. The date-based ranking methods follow the general hypothesis that more information on a subject becomes gradually available over time and newer evidence should therefore be ranked higher than older evidence \citep{li2003time}. Both \textit{evidence-based date ranking} and \textit{claim-based date ranking} rank evidence by date in descending order, but the latter excludes evidence that is published after the claim date. While these date-based ranking methods consider relevance as a property proportional to recency, the distance-based ranking methods define relevance in terms of temporal distance to the claim (\textit{claim-centered distance ranking}) or the other evidence in the evidence set (\textit{evidence-centered distance ranking}). In these ranking methods, evidence is ranked by distance in ascending order. Table \ref{tab:methods_overview} provides an overview of all methods illustrated with simulated ground-truth rankings for sample evidence set \(E_s\). 

The dataset used for training and testing contains claims from multiple fact-check domains. In Section \ref{results} (Results), we take the model performance results for each fact-check domain when applying one of the temporal ranking methods and average over all domains to report the overall impact of each ranking method. We also report \textit{time-aware evidence ranking} performance by taking the best performing temporal ranking method for each domain and averaging over all domains. This way, we show how choosing the appropriate ranking method for each domain influences model performance.

\subsubsection{Evidence-based date ranking}

\begin{theorem}
Evidence published later in time is more relevant than evidence published earlier in time and should thus be ranked accordingly.
\end{theorem}
The evidence-based date ranking method ranks the evidence snippets in a given set by their publication date in descending order. The simulated ground-truth ranking \(R_i =  \{r(e_{i_{1}}),...,r(e_{i_{K}})\}\) 
satisfies the following constraint:
\begin{equation}
\begin{split}
    & \forall e_{i_{j}},e_{i_{k}} \in E_i : {\Delta}t(e_{i_{j}}) \leq {\Delta}t(e_{i_{k}}) \\ &\implies r(e_{i_{j}}) \leq r(e_{i_{k}})
\end{split}
\end{equation}
For two evidence snippets in the same evidence set, the constraint imposes a higher ranking score \(r(e_{i_{k}})\) for \(e_{i_{k}}\) and a lower ranking score \(r(e_{i_{j}})\) for \(e_{i_{j}}\) if \({\Delta}t(e_{i_{k}})\) is larger than \({\Delta}t(e_{i_{j}})\). If \({\Delta}t(e_{i_{k}})\) and \({\Delta}t(e_{i_{j}})\) are equal, \(e_{i_{k}}\) and \(e_{i_{j}}\) obtain the same ranking score. Following the intuition that recent evidence is more relevant than past evidence, an evidence snippet that is published more recently is assigned a higher position in the evidence ranking than an evidence snippet that is published further in the past.
\\ \\
\textbf{Example}: For \(E_s = \{e_{s_1}, e_{s_2}, e_{s_3}, e_{s_4}\}\) with \({\Delta}E^t_s = \{{\minus}3, 0, {\minus}8791, 95\}\), the simulated ground-truth ranking \(R_s = \{r(e_{s_1}),r(e_{s_2}),r(e_{s_3}),r(e_{s_4})\}\) should satisfy the method-specific constraint (7): 
\(\forall e_{s_{j}} \in E_s : {(\Delta}t(e_{s_{3}}) = \minus8791)<{(\Delta}t(e_{s_{1}})=\minus3)<({\Delta}t(e_{s_{2}})=0)< ({\Delta}t(e_{s_{4}})=95) \implies (r(e_{s_3})=1)<(r(e_{s_1})=2)<(r(e_{s_2})=3)<(r(e_{s_4})=4)\). Hence, \(R_s = \{2,3,1,4\}\).

\subsubsection{Claim-based date ranking}
\begin{theorem}
Although newer evidence is more relevant than older evidence, fact-checkers can only base their veracity estimation on information that was available at fact checking time. In this case, evidence published after the claim date is not relevant and should be excluded from the documents to be ranked.
\end{theorem}
In this method, we mimic the information accessibility and availability at claim time \(t(c_i)\). As a result, only evidence that had been published before or at the same time as the claim is considered in this approach. For ground-truth evidence ranking \(R_i\),  
all \(e_{i_{j}}\) with 
\({\Delta}t(e_{i_{j}}) \leq 0\) are ranked according to the following constraint:
\begin{equation}
    \begin{split}
    & \forall e_{i_{j}}, e_{i_{k}} \in E_i:\\ & ({\Delta}t(e_{i_{j}}) \leq {\Delta}t(e_{i_{k}})) \land ({\Delta}t(e_{i_{j}}) \land {\Delta}t(e_{i_{k}}) \leq 0) \\ & \implies r(e_{i_{j}}) \leq r(e_{i_{k}})
    \end{split}
\end{equation}
For two evidence snippets in the same evidence set, the constraint imposes a higher ranking score \(r(e_{i_{k}})\) for \(e_{i_{k}}\) and a lower ranking score \(r(e_{i_{j}})\) for \(e_{i_{j}}\) if \({\Delta}t(e_{i_{k}})\) is larger than \({\Delta}t(e_{i_{j}})\) and both \({\Delta}t(e_{i_{k}})\) and \({\Delta}t(e_{i_{j}})\) are negative or zero. If \({\Delta}t(e_{i_{k}})\) and \({\Delta}t(e_{i_{j}})\) are equal, \(e_{i_{j}}\) and \(e_{i_{k}}\) obtain the same ranking score. This is the only temporal ranking method that does not necessarily rank all given \(e_{i_j} \in E_i\) because it excludes all \({\Delta}t(e_{i_j}) > 0\) from simulated ground-truth ranking \(R_i\). Following the intuitions that recent evidence is more relevant than past evidence and fact-checkers can only rely on information that was available at claim time, an evidence snippet that is published more recently and before or at claim time is assigned a higher position in the evidence ranking than an evidence snippet that is published further in the past. 
\\ \\
\textbf{Example}: For \(E_s = \{e_{s_1}, e_{s_2}, e_{s_3}, e_{s_4}\}\) with \({\Delta}E^t_s = \{{\minus}3, 0, {\minus}8791, 95\}\), the simulated ground-truth ranking \(R_s = \{r(e_{s_1}),r(e_{s_2}),r(e_{s_3}),r(e_{s_4})\}\) should satisfy the method-specific constraint (8): 
\(\forall e_{s_{j}} \in E_s : (({\Delta}t(e_{s_{3}}) = \minus8791)<({\Delta}t(e_{s_{1}})=\minus3)<({\Delta}t(e_{s_{2}})=0)) \land ({\Delta}t(e_{s_{1}})\land{\Delta}t(e_{s_{2}})\land{\Delta}t(e_{s_{3}}) \leq 0) \implies (r(e_{s_3})=1)<(r(e_{s_1})=2)<(r(e_{s_2})=3)\). The ranking scores for \({\Delta}t(e_{s_j})>0\) are automatically set to 0. Hence, \(R_s = \{2,3,1,0\}\). 

\subsubsection{Claim-centered distance ranking}
\begin{theorem}
Assuming that a topic and its related subtopics are discussed around the same time \citep{martins2019modeling}, evidence is more relevant when it is published around the same time as the claim and becomes less relevant as the temporal distance between claim and evidence grows.
\end{theorem}
This ranking method assigns ranking scores to evidence snippets in terms of their temporal vicinity to the claim. 
For ground truth ranking \(R_i\),
we rank all \(e_{i_{j}}\) based on \(\abs{{\Delta}t(e_{i_{j}})}\) in ascending order, satisfying the following constraint:
\begin{equation}
    \begin{split}
    & \forall e_{i_{j}}, e_{i_{k}} \in E_i : \abs{{\Delta}t(e_{i_{j}})} \geq \abs{{\Delta}t(e_{i_{k}})} \\ & \implies r(e_{i_{j}}) \leq r(e_{i_{k}})
    \end{split}
\end{equation}
For two evidence snippets in the same evidence set, the constraint imposes a higher ranking score \(r(e_{i_{j}})\) for \(e_{i_{j}}\) and a lower ranking score \(r(e_{i_{k}})\) for \(e_{i_{k}}\) if \(\abs{{\Delta}t(e_{i_{j}})}\) is smaller than \({\abs{{\Delta}t(e_{i_{k}})}}\). If \({\abs{{\Delta}t(e_{i_{j}})}}\) and \({\abs{{\Delta}t(e_{i_{k}})}}\) are equal, \(e_{i_{j}}\) and \(e_{i_{k}}\) obtain the same ranking score. Following the intuition that relevant evidence clusters around the claim in terms of time, an evidence snippet with a timestamp close to claim time is assigned a higher position in the evidence ranking than an evidence snippet with a timestamp distant from claim time.
\\ \\
\textbf{Example}: For \(E_s = \{e_{s_1}, e_{s_2}, e_{s_3}, e_{s_4}\}\) with \({\Delta}E^t_s = \{{\minus}3, 0, {\minus}8791, 95\}\), the simulated ground-truth ranking \(R_s = \{r(e_{s_1}),r(e_{s_2}),r(e_{s_3}),r(e_{s_4})\}\) should satisfy the method-specific constraint (9): 
\(\forall e_{s_{j}} \in E_s : (|{\Delta}t(e_{s_{3}})| = 8791)>(|{\Delta}t(e_{s_{4}})|=95)>(|{\Delta}t(e_{s_{1}})|=3)>(|{\Delta}t(e_{s_{2}})|=0) \implies (r(e_{s_3})=1)<(r(e_{s_4})=2)<(r(e_{s_1})=3)<(r(e_{s_2})=4)\). Hence, \(R_s = \{3,4,1,2\}\).

\subsubsection{Evidence-centered distance ranking}
\begin{theorem}
Analogous to the assumption that relevant documents have a tendency to cluster in a shared document space \citep{efron2014time}, relevant evidence snippets also cluster in time. Therefore, evidence snippets that are clustered in time are more relevant than evidence snippets that are temporally distant from the others.
\end{theorem}
This ranking method assigns ranking scores to evidence snippets in terms of their temporal vicinity to the other snippets in the same evidence set. We first detect the medoid of all \({\Delta}t(e_{i_{j}}) \in {\Delta}E_i^t\) by computing a pairwise distance matrix, summing the columns and finding the argmin of the summed pairwise distance values.
Then, the Euclidean distance between each \({\Delta}t(e_{i_{j}})\) and the detected medoid is calculated. 
\begin{equation}
    {\Delta}cl: {\Delta}E^t_i \xrightarrow{} {\Delta}E_i^{cl} = \{{\Delta}cl(e_{i_{j}}),...,{\Delta}cl(e_{i_{K}})\}
\end{equation}
We rank all \( {\Delta}cl(e_{i_j}) \in {\Delta}E_i^{cl}\) in ascending order, resulting in ground-truth ranking \(R_i\) which satisfies the following constraint:
\begin{equation}
    \begin{split}
    & \forall e_{i_{j}}, e_{i_{k}} \in E_i : \abs{{\Delta}cl(e_{i_{j}})} \geq \abs{{\Delta}cl(e_{i_{k}})} \\ & \implies r(e_{i_{j}}) \leq r(e_{i_{k}})
    \end{split}
\end{equation}
For two evidence snippets in the same evidence set, the constraint imposes a higher ranking score \(r(e_{i_{j}})\) for \(e_{i_{j}}\) and a lower ranking score \(r(e_{i_{k}})\) for \(e_{i_{k}}\) if \(\abs{{\Delta}cl(e_{i_{j}})}\) is smaller than \({\abs{{\Delta}cl(e_{i_{k}})}}\). If \({\abs{{\Delta}cl(e_{i_{j}})}}\) and \({\abs{{\Delta}cl(e_{i_{k}})}}\) are equal, \(e_{i_{j}}\) and \(e_{i_{k}}\) obtain the same ranking score. Following the intuition that relevant evidence clusters in time, an evidence snippet with a timestamp close to the evidence mediod timestamp is assigned a higher position in the evidence ranking than an evidence snippet with a timestamp distant from the evidence mediod timestamp.
\\ \\
\textbf{Example}: In order to obtain \({\Delta}E^{cl}_s = \{{\Delta}cl(e_{s_1}), {\Delta}cl(e_{s_2}), {\Delta}cl(e_{s_3}), {\Delta}cl(e_{s_4})\}\): 
\begin{itemize}
    \item Compute the pairwise distance matrix for \({\Delta}E^t_s = \{\minus3,0,\minus8791,95\}\): \\
    $ PD =
    \begin{Bmatrix}
        0 & 3 & 8788 & 98 \\
        3 & 0 & 8791 & 95 \\
        8788 & 8791 & 0 & 8886 \\
        98 & 95 & 8886 & 0 \\
    \end{Bmatrix}
    $
    \item Sum the matrix columns: \\
    $ SD = 
    \begin{Bmatrix}
        8889 & 8889 & 26465 & 9079 \\
    \end{Bmatrix}
    $
    \item Find the argmin and mediod\footnote{In case of ties when computing the mediod, we select the first one in the evidence set. For example, if the algorithm would find \(e_{s_1}\) and \(e_{s_2}\) as mediods, \(e_{s_1}\) is selected.}: \\ $argmin(SD) = 8889$, \(mediod = {\Delta}t(e_{s_1}) = \minus3\) 
    \item Calculate the Euclidean distance between all \({\Delta}t(e_{s_j})\) and the detected mediod: \\
    \({\Delta}cl: {\Delta}E^t_s \xrightarrow{} {\Delta}E_s^{cl} = \{0,3,8788,98\}\)
\end{itemize}

For \(E_s = \{e_{s_1}, e_{s_2}, e_{s_3}, e_{s_4}\}\) with \({\Delta}E_s^{cl} = \{0,3,8788,98\}\), the simulated ground-truth ranking \(R_s = \{r(e_{s_1}),r(e_{s_2}),r(e_{s_3}),r(e_{s_4})\}\) should satisfy the method-specific constraint (11): 
\(\forall e_{s_{j}} \in E_s : (|{\Delta}cl(e_{s_{3}})| = 8788) >(|{\Delta}cl(e_{s_{4}})|=98)>(|{\Delta}cl(e_{s_{2}})|=3)>(|{\Delta}cl(e_{s_{1}})|=0) \implies (r(e_{s_3})=1)<(r(e_{s_4})=2)<(r(e_{s_2})=3)<(r(e_{s_1})=4)\). Hence, \(R_s = \{4,3,1,2\}\).

%% file: model_architecture.tex
\subsection{Fact-Checking Model Architecture}

\begin{figure*}
  \begin{subfigure}{\linewidth}
    \centering
    \includegraphics[width=12cm]{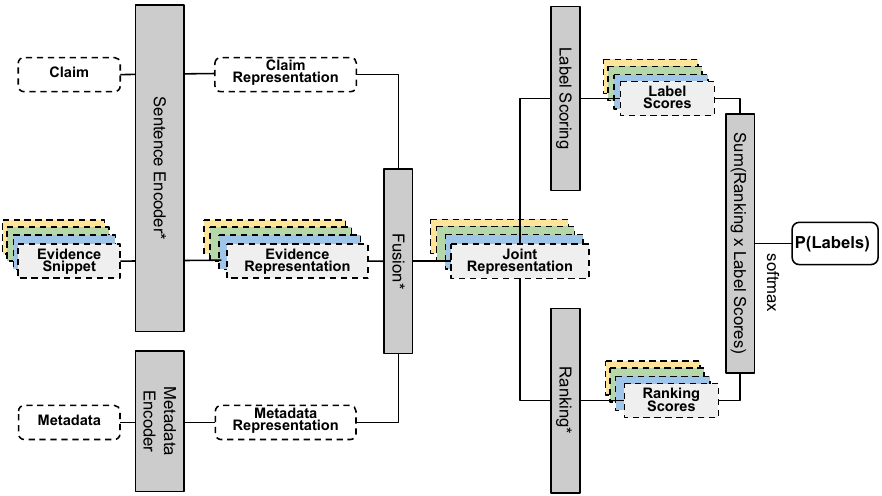}
    \caption{BiLSTM/RNN}
    \label{fig:rnn}
  \end{subfigure}
  \begin{subfigure}{\linewidth}
    \centering
    \includegraphics[width=12cm]{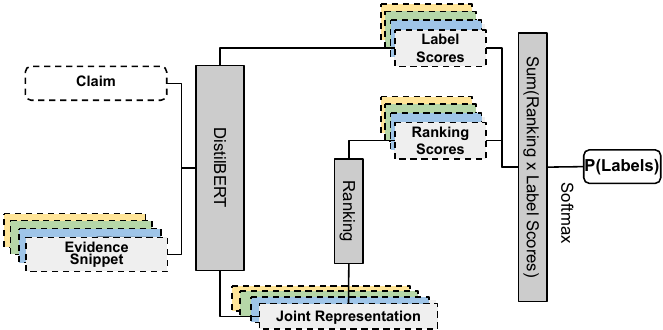}
    \caption{DistilBERT}
    \label{fig:bert}
  \end{subfigure}
  \caption{Overview of the fact-checking models. The BiLSTM and RNN model (a) share a similar model structure but differ in sentence encoder architecture (BiLSTM vs. RNN), fusion mechanism (inference matching vs. concatenation) and number of fully-connected layers in the ranking module (two vs. one). The colored boxes with dashed borders indicate vectors while the grey boxes with solid borders indicate model modules. Each color refers to an evidence snippet, showing how the evidence set is encoded throughout the model.}
  \label{fig:main_model}
\end{figure*}

In this section, we describe the model architecture of three fact-checking models. 
In a training setup without any ranking constraints, all model parameters are optimized using a loss function on the verification classification task. Evidence ranking is then learned implicitly. 
In a training setup with our ranking constraints, a model's ranking parameters are directly optimized using a learning-to-rank loss on the yielded evidence rankings while the other model parameters are optimized using a loss on the predicted veracity labels. 
By applying our temporal ranking methods to various neural architectures, we show their advantage over time-unaware approaches in a transparent manner. We take the Joint Veracity Prediction and Evidence Ranking model presented in the MultiFC dataset paper \citep{augenstein2019multifc} as base model architecture and experiment with different neural architectures for the other two models. For the sake of simplicity, we name the models by their sentence encoder architecture. An overview of the model architectures is given in Figure \ref{fig:main_model}.

\subsubsection{BiLSTM}
This is the Joint Veracity Prediction and Evidence Ranking model as presented in the dataset paper \citep{augenstein2019multifc}.
The BiLSTM model takes as input claim sentence \(c_i\), evidence set \(E_i = \{e_{i_{1}},...,e_{i_{K}}\}\) and claim metadata\footnote{Metadata contains information on speaker, tags and categories.} \(m_i\) - with \(K \leq 10\)
the total number of evidence snippets in evidence set \(E_i\). The sentence encoder - a bidirectional LSTM with skip-connections - transforms \(c_i\) and \(e_{i_j}\) into their respective hidden representation \(h_{c_i}\) and \(h_{e_{i_j}}\) while the metadata encoder - a CNN - transforms \(m_i\) into \(h_{m_i}\).
Each \(h_{e_{i_{j}}}\) is then fused with \(h_{c_{i}}\) and \(h_{m_{i}}\) into a joint representation \(f_{i_j}\) following a natural language inference matching method proposed by \citet{DBLP:conf/acl/MouMLX0YJ16}: 
\begin{equation}\label{fusion}
    f_{i_{j}} = [h_{c_{i}};h_{e_{i_{j}}};h_{c_{i}}-h_{e_{i_{j}}};h_{c_{i}}{\circ}h_{e_{i_{j}}};h_{m_{i}}]
\end{equation}
where semi-colon denotes vector concatenation, ``\(-\)" element-wise difference and ``\(\circ\)" element-wise multiplication.
All \(f_{i_j}\) are sent through two modules: a label scoring and a ranking module. In the label scoring module, similarity between \(f_{i_{j}}\) and each label 
across all domains is scored by taking the dot product between \(f_{i_j}\) and all label embeddings, which are updated during model training. This way, relationships between labels across all fact-check domains are learned. This results in label similarity matrix \(S_{i_{j}}\). As the model needs to predict a domain-specific label, a domain-specific mask is applied over \(S_{i_{j}}\), masking all out-of-domain label similarity scores. Ultimately, a fully-connected layer computes domain-specific label score vector \(l_{i_{j}}\). In the ranking module, a two-layer fully-connected layer computes a ranking score \(r(f_{i_j})\) for each \(f_{i_j}\). The ranking score reflects the relevance of \(e_{i_j}\): the higher the score, the higher the evidence snippet's relevance to the claim. 
The dot product between each label score vector \(l_{i_{j}}\) and its respective ranking score \(r(f_{i_{j}})\) is taken, the resulting vectors are summed and domain-specific label probabilities \(p_i\) are obtained by applying a softmax function. Finally, the model outputs the domain-specific veracity label with the highest probability.

\subsubsection{RNN}

This model's architecture is similar to that of the BiLSTM model but differs in terms of sentence encoder architecture, fusion mechanism and number of fully-connected layers in the ranking module. Instead of a bidirectional LSTM with skip-connections, a two-layer unidirectional RNN encodes \(c_i\) and \(e_{i_j}\in E_i\) into their hidden representations \(h_{c_i}\) and \(h_{e_{i_j}}\). These representations are then fused with the hidden metadata representation \(h_{m_i}\) using a simple concatenation operation instead of the natural language inference matching method: \(f_{i_j} = [h_{c_i};h_{e_{i_j}};h_{m_i}]\). Lastly, we halve the number of fully-connected layers in the ranking module so that the ranking module is a shallower network. 

\subsubsection{DistilBERT}

In this model, a DistilBERT Transformer model with a sequence classification head on top\footnote{We adopt the pre-trained DistilBertForSequenceClassification model from the Hugging Face Transformers library \cite{sanh2019distilbert}.} jointly takes as input claim sentence \(c_i\) and evidence snippet \(e_{i_j}\). It returns the `[CLS]' embedding from its final contextual layer and a probability distribution over all labels \(p_{i_j}\). The `[CLS]' embedding represents the joint claim-evidence representation \(f_{i_j}\). All joint representations \(f_{i_j}\) are then sent through a two-layer fully-connected layer that computes a ranking score \(r(f_{i_j})\) for each \(f_{i_j}\). As DistilBERT already yields a probability distribution over all labels \(p_{i_j}\), we simply apply a domain-specific mask over \(p_{i_j}\) to obtain label score vector \(l_{i_j}\). The dot product between each label score vector \(l_{i_{j}}\) and its respective ranking score \(r(f_{i_{j}})\) is taken, the resulting vectors are summed and domain-specific label probabilities \(p_i\) are obtained by applying a softmax function. Finally, the model outputs the domain-specific veracity label with the highest probability.

%% file: experiments.tex
\section{Experimental Setup}

\textbf{Dataset}. We opt for the MultiFC dataset \citep{augenstein2019multifc} as it is the only large, publicly available fact-checking dataset which provides temporal information for both claims and evidence pages and follow their experimental setup. The dataset contains 34,924 real-world claims extracted from 26 different fact-check websites. The fact-check domains are abbreviated to four-letter contractions (e.g. Snopes to \textit{snes}, Washington Post to \textit{wast}). For each claim, metadata such as speaker, tags and categories are included, and a maximum of ten evidence snippets crawled from the Internet using the Google Search API are used to predict a claim's veracity label. 
To retrieve evidence snippets, \cite{augenstein2019multifc} submit each claim verbatim as a query without quotes.

\begin{figure}
    \centering
    \includegraphics[width=\linewidth]{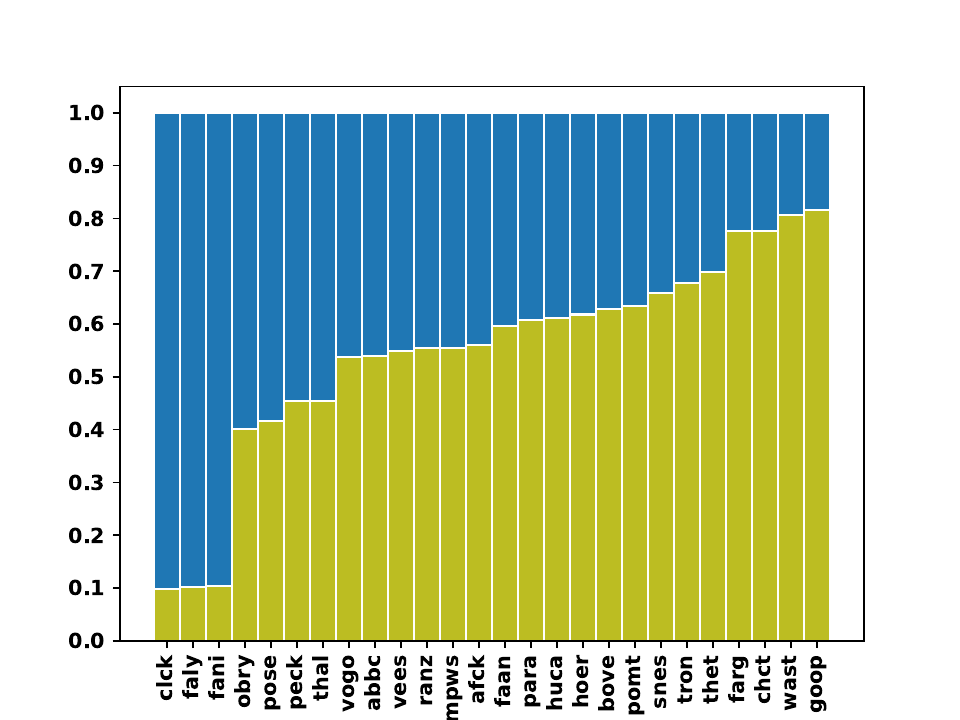}
    \caption{The share of evidence snippets with retrievable timestamps per domain (in yellow).}
    \label{fig:time_distribution}
\end{figure}

Regarding temporal information, the dataset provides an explicit timestamp for each claim as structured metadata. For the evidence snippets, however, we need to extract their timestamp from the evidence text itself. The publication date is often contained in the document text, most frequently at the beginning of the text and immediately followed by an ellipsis (i.e., `...'). We split the text at the ellipsis and take the left part as timestamp. If Python's datetime module recognizes the extracted timestamp as a real timestamp, it is regarded as ground-truth evidence timestamp. Otherwise, the timestamp is not included in the dataset. 
Both claim and evidence date are then automatically formatted as \textit{year-month-day} using Python's datetime module. We randomly extracted 150 timestamps from claims and evidence snippets in the dataset and manually verified whether datetime correctly parsed them in \textit{year-month-day}. As zero mistakes were found, we consider datetime sufficiently accurate. If datetime is unable to correctly format an extracted timestamp, we do not include it in the dataset. 
Figure \ref{fig:time_distribution} displays the distribution of evidence snippets with and without a retrievable timestamp per domain. 
Finally, the dataset is split in training (27,940), development (3,493) and test (3,491) set in a label-stratified manner.
\\ \\
\textbf{Pre-Training and Fine-Tuning}. The models are first pre-trained on all domains before they are fine-tuned for each domain separately. Pre-training the models on all domains is advantageous as some domains contain few training data. In the pre-training phase, batches from each domain are alternately fed to the model during each epoch with the maximum number of batches for all domains equal to the maximum number of batches for the smallest domain. This way, the model is not biased towards the larger domains. For the veracity classification task, cross-entropy loss is computed over the label probabilities, and RMSProp (BiLSTM/RNN) or AdamW (DistilBERT) optimizes all the model parameters except the evidence ranking parameters. For the evidence ranking task, ListMLE loss is computed over the ranking scores, and Adam optimizes only the evidence ranking parameters. We use the ListMLE loss function from the allRank library \cite{pobrotyn2020ContextAwareLT}. An overview of all hyperparameter settings is included in the Appendix. 
In the fine-tuning phase, we select the best performing pre-trained model based on the development set for each domain individually and fine-tune it on that domain.

We found that directly optimizing the evidence ranking on the temporal ranking methods in both the pre-training and fine-tuning phase yields the highest results for all models (Table \ref{tab:pt_or_ft}). 

\begin{table}
    \small
    \centering
    \begin{tabular}{cccc}
    \toprule
        Phase & BiLSTM & RNN & DistilBERT \\
        \midrule
        Only PT & .5998/.3436 & .5425/.2960 & .5891/\textbf{.3282} \\
        Only FT & .6072/.3525 & .5862/\textbf{.3513} & .5424/.2859 \\
        PT + FT & \textbf{.6265}/\textbf{.3673} & \textbf{.5899}/.3453 & \textbf{.5921}/.3135 \\
    \bottomrule
    \end{tabular}
    \caption{Overview of the aggregated test results (Micro/Macro F1) when optimized on the temporal ranking methods. The methods are integrated during pre-training only (Only PT), during fine-tuning only (Only FT) or during both pre-training and fine-tuning (PT + FT).}
    \label{tab:pt_or_ft}
\end{table}



%% file: results.tex
\section{Results}\label{results}

\begin{table*}
    \fontsize{8.5}{9}\selectfont
    \setlength{\tabcolsep}{2pt}
    \centering
    \begin{tabular}{ccc|cc|cc}
        \toprule
         & \multicolumn{2}{c}{\textbf{BiLSTM}} & \multicolumn{2}{c}{\textbf{RNN}} & \multicolumn{2}{c}{\textbf{DistilBERT}} \\
        \cmidrule{2-7} 
        & Micro F1 & Macro F1 & Micro F1 & Macro F1 & Micro F1 & Macro F1  \\
        \midrule
        \textbf{Base model} & 
        \textbf{.5521} & \textbf{.3185} & \textbf{.5053} & \textbf{.2691} & \textbf{.5558} & \textbf{.2909}\\
        \midrule
        Random evidence & .5383$_{\minus1.38\%}$ & .2919$_{\minus2.66\%}$ & \underline{.5196}$_{+1.43\%}$ & \underline{.2818}$_{+1.27\%}$ & .5040$_{\minus5.18\%}$ & .2495$_{\minus4.14\%}$ \\
        \midrule
        Search ranking & .5468$_{\minus0.53\%}$ & .2864$_{\minus3.21\%}$ & .4782$_{\minus2.71\%}$ & .2372$_{\minus3.19\%}$ & .4793$_{\minus7.65\%}$ & .2264$_{\minus6.45\%}$ \\
        \midrule
        Evidence-based date ranking & \underline{.5794}$_{+2.73\%}$ & \underline{.3227}$_{+0.42\%}$ & \underline{.5359}$_{+3.06\%}$ & \underline{.2905}$_{+2.14\%}$ & .5424$_{\minus1.34\%}$ & .2660$_{\minus2.49\%}$ \\
        Claim-based date ranking & \underline{.5668}$_{+1.47\%}$ & \underline{.3215}$_{+0.30\%}$ & \underline{.5271}$_{\minus2.18\%}$ & \underline{.3082}$_{+3.91\%}$ & .4791$_{\minus7.67\%}$ & .2391$_{\minus5.18\%}$ \\
        Claim-centered distance ranking & \underline{.5532}$_{+0.11\%}$ & .3029$_{\minus1.56\%}$ & .4981$_{\minus0.72\%}$ & \underline{.2700}$_{\minus0.09\%}$ & .5257$_{\minus3.01\%}$ & .2644$_{\minus2.65\%}$ \\
        Evidence-centered distance ranking & .4756$_{\minus7.65\%}$ & .2417$_{\minus7.68\%}$ & \underline{.5356}$_{+3.03\%}$ & \underline{.3019}$_{3.28\%}$ & \underline{.5759}$_{+2.01\%}$ & \underline{.2962}$_{+0.53\%}$ \\
        \midrule
        \textbf{Time-aware evidence ranking} & \textbf{.6265}$_{+7.44\%}$ & \textbf{.3673}$_{+4.88\%}$ & \textbf{.5899}$_{+8.46\%}$ & \textbf{.3453}$_{+7.62\%}$ & \textbf{.5921}$_{+3.63\%}$ & \textbf{.3135}$_{+2.26\%}$ \\
        \bottomrule
    \end{tabular}
    \caption{Aggregated test results for the veracity prediction task, with improvements over base model performance underlined and the performances differences as subscript.}
    \label{tab:results}
\end{table*}

We take the best performing model per domain based on the development set and report the average over all domain-specific test results on the veracity prediction task (Table \ref{tab:results}). As done in \citet{augenstein2019multifc}, we use Micro and Macro F1 score as evaluation metrics\footnote{We use the \texttt{f1\_score} metric from the scikit-learn library in Python to compute the Micro and Macro F1 score.}. The results of our BiLSTM base model are comparable\footnote{\citet{augenstein2019multifc} train the model differently: they train the multi-task learning model over all domains and perform early stopping based on accuracy on the target domain development set. We pre-train the model over all domains, select the best performing model per domain based on the development set, and fine-tune each domain-specific model individually. Hence the difference in test results.} to those of \citet{augenstein2019multifc}. In the \textit{random evidence} experiment, we establish how well the models distinguish between the relevance of extracted evidence sets that are considered relevant by the Google Search API and the assumed irrelevance of evidence sets that are randomly assigned to the claims (random seed = 0). As an additional ranking baseline, we optimize evidence ranking on the evidence snippets' position in the Google search ranking list (\textit{search ranking}). As the MultiFC dataset kept the order in which the Google Search API provided the first ten query results for each claim query, we can simply take that order for the \textit{search ranking} setup: the higher in the provided evidence set, the higher in the simulated ground-truth evidence ranking.
For the \textit{time-aware evidence ranking} results, we take for each domain the best performing temporal ranking method and report again the average over all domain-specific test results.
We present domain-specific test results for the BiLSTM model in Table \ref{tab:results_multifc} and include the domain-specific test results for the RNN and DistilBERT model in the Appendix. Given that the number of classification labels per domain range from 2 to 40 labels, we discuss the performance results in terms of Micro F1 as we suspect label imbalance.

\begin{table*}[!htpb]
    \fontsize{7.5}{8.5}\selectfont
    \setlength{\tabcolsep}{1.5pt}
    \centering
    \begin{tabular}{ccc||cc||cccccccc||cc}
        \toprule  & 
        \multicolumn{2}{c}{\textbf{Base}} & \multicolumn{2}{c}{\textbf{Search}} & \multicolumn{2}{c}{\textbf{Evidence}} & \multicolumn{2}{c}{\textbf{Claim}} &
        \multicolumn{2}{c}{\textbf{Claim}} &
        \multicolumn{2}{c}{\textbf{Evidence}} &
        \multicolumn{2}{c}{\textbf{Time-Aware}}\\
        & \multicolumn{2}{c}{\textbf{Model}} & \multicolumn{2}{c}{\textbf{Ranking}} & \multicolumn{2}{c}{\textbf{Date}} & \multicolumn{2}{c}{\textbf{Date}} &
        \multicolumn{2}{c}{\textbf{Distance}} &
        \multicolumn{2}{c}{\textbf{Distance}} &
        \multicolumn{2}{c}{\textbf{Ranking}}\\
        \cmidrule{2-15}
         & Micro & Macro & Micro & Macro & Micro & Macro & Micro & Macro & Micro & Macro & Micro & Macro & Micro & Macro \\
         \midrule
        abbc & \textbf{.3148} & \textbf{.2782} & .5370 & .2329 & .5000 & .2222 & .4630 & .2137 & .4074 & .2578 & .5370 & .2358 & \textbf{\underline{.5370}} & \textbf{\underline{.2358}} \\
        afck & \textbf{.2222} & \textbf{.1385} & .1667 & .0706 & .2778 & .0725 & .2778 & .0741 & .2778 & .0725 & .2222 & .0571  & \textbf{\underline{.2778}} & \textbf{\underline{.0741}} \\
        bove & \textbf{1.} & \textbf{1.} & 1. &  1. & 1. & 1. & 1. & 1. & 1. & 1. & 1. & 1. & \textbf{1.} & \textbf{1.} \\
        chct & \textbf{.5500} & \textbf{.2967} & .4250 & .2294 & .4750 & .2956 & .4000 & .2234 & .4500 & .1552 & .4000 & .2267 & \textbf{.4750} & \textbf{.1959} \\
        clck & \textbf{.5000} & \textbf{.2500} & .1667 & .0952 & .8333 & .6061 & .6667 & .5333 & .5000 & .2500 & .5000 & .2500 & \textbf{\underline{.8333}} & \textbf{\underline{.6061}} \\
        faan & \textbf{.5000} & \textbf{.5532} & .1364 & .1216 & .5455 & .4545 & .5000 & .3266 & .5000 & .4256 & .4091 & .2837 & \textbf{\underline{.5455}} & \textbf{\underline{.4545}} \\
        faly & \textbf{.6429} & \textbf{.2045} & .6429 & .1957 & .8571 & .4521 & .6429 & .1957 & .4286 & .1667 & .2857 & .3000 & \textbf{\underline{.8571}} & \textbf{\underline{.4521}} \\
        fani & \textbf{1.} & \textbf{1.} & 1. & 1.  & 1. & 1. & 1. & 1. & 1. & 1. & 1. & 1. & \textbf{1.} & \textbf{1.} \\
        farg & \textbf{.6923} & \textbf{.1023} & .6923 & .1023 & .6923 & .1023 & .6731 & .1006 & .6923 & .1023 & .7308 & .2003 & \textbf{\underline{.7308}} & \textbf{\underline{.2003}} \\
        goop & \textbf{.8344} & \textbf{.1516} & .8344 & .1516 & .8344 & .6222 & .8245 & .1512 & .8278 & .1512 & .8344 & .1516 & \textbf{.8344} & \textbf{.1516} \\
        hoer & \textbf{.4104} & \textbf{.2091} & .4403 & .1975 & .4627 & .2200 & .3955 & .1474 & .4701 & .2854 & .3209 & .1106 & \textbf{\underline{.4701}} & \textbf{\underline{.2854}} \\
        huca & \textbf{.5000} & \textbf{.2857} & .5000 & .2857 & .8333 & .6222 & .5000 & .2857 & .6667 & .5333 & .6667 & .5926 & \textbf{\underline{.8333}} & \textbf{\underline{.6222}} \\
        mpws & \textbf{.8750} & \textbf{.6316} & .8750 & .6316 & .5000 & .2222 & .8750 & .6316 & .5000 & .2424 & .2500 & .1333 & \textbf{.8750} & \textbf{.6316} \\
        obry & \textbf{.5714} & \textbf{.3520} & .3571 & .1316 & .3571 & .1316 & .4286 & .1500 & .5000 & .2982 & .3571 & .1316 & \textbf{.5000} & \textbf{.2982} \\
        para & \textbf{.1875} & \textbf{.1327} & .2500 & .0667 & .3125 & .1720 & .2813 & .1584 & .1563 & .0490 & .1875 & .0556 & \textbf{\underline{.3125}} & \textbf{\underline{.1720}} \\
        peck & \textbf{.5833} & \textbf{.2593} & .9167 & .6154 & .5833 & .2593 & .6667 & .8213 & .9167 & .6316 & .2500 & .1724 & \textbf{\underline{.9167}} & \textbf{\underline{.6316}} \\
        pomt & \textbf{.2143} & \textbf{.2089} & .2071 & .1015 & .1732 & .0604 & .1643 & .0539 & .1732 & .0329 & .1813 & .0594 & \textbf{.1813} & \textbf{.0594} \\
        pose & \textbf{.4178} & \textbf{.0987} & .4178 & .0987 & .3836 & .1112 & .4247 & .1330 & .4041 & .1990 & .4110 & .1422 & \textbf{\underline{.4247}} & \textbf{\underline{.1330}} \\
        ranz & \textbf{1.} & \textbf{1.} & 1. & 1. & 1. & 1. & 1. & 1. & 1. & 1. & .5000 & .3333 & \textbf{1.} & \textbf{1.} \\
        snes & \textbf{.5646} & \textbf{.0601} & .5646 & .0614 & .5662 & .1026 & .5646 & .0753 & .5646 & .0601 & .5646 & .0602 & \textbf{\underline{.5662}} & \textbf{\underline{.1026}} \\
        thal & \textbf{.5556} & \textbf{.2756} & .5556 & .1786 & .4444 & .1538 & .5556 & .1852 & .5556 & .1852 & .5556 & .1852 & \textbf{.5556} & \textbf{.1852} \\
        thet & \textbf{.3750} & \textbf{.2593} & .6250 & .3556  & .6250 & .3556 & .6250 & .3556 & .6250 & .3556 & .4375 & .1014 & \textbf{\underline{.6250}} & \textbf{\underline{.3556}} \\
        tron & \textbf{.4767} & \textbf{.0258} & .4767 & .0325 & .4738 & .0258 & .4767 & .0258 & .4622 & .0289 & .4767 & .0258 & \textbf{.4767} & \textbf{\underline{.0366}} \\
        vees & \textbf{.7097} & \textbf{.2075} & .7097 & .2075 & .6613 & .1990 & .6129 & .2179 & .7097 & .2075  & .6935 & .2885 & \textbf{.7097} & \textbf{.2075} \\
        vogo & \textbf{.5000} & \textbf{.2056} & .5000 & .2158 & .4219 & .2531 & .5000 & .2262 & .3750 & .1234 & .3750 & .1114 & \textbf{.5000} & \textbf{\underline{.2262}} \\
        wast & \textbf{.1563} & \textbf{.0935} & .2188 & .0667 & .2500 & .1429 & .2188 & .0718 & .2188 & .1412 & .2188 & .0757 & \textbf{\underline{.2500}} & \textbf{\underline{.1429}} \\
        \midrule
        avg. & \textbf{.5521} & \textbf{.3185} & .5468 & .2864 & \underline{.5794} & \underline{.3227} & \underline{.5668} & \underline{.3215} & \underline{.5532} & .3029  & .4756 & .2417 & \textbf{\underline{.6265}} & \textbf{\underline{.3673}} \\
        \bottomrule
    \end{tabular}
    \caption{Overview of classification test results (Micro F1 and Macro F1) for \textbf{BiLSTM model - PT + FT}, with improvements over base model results \underline{underlined}. The last column contains the highest results returned by one of the four temporal ranking methods.}
    \label{tab:results_multifc}
\end{table*}

The \textit{random evidence} results indicate that the DistilBERT and BiLSTM base models are able to distinguish relevant evidence sets from random evidence sets without any guidance on evidence ranking (-1.38\%/-5.18\% Micro F1; BiLSTM/DistilBERT).
Regarding the temporal ranking methods, the methods generally outperform search engine ranking, which consistently performs worse than the base models (-0.53/-2.71/-7.65\% Micro F1; BiLSTM/RNN/DistilBERT). The temporal ranking methods seem to affect model performance to various extents. Constraining evidence ranking in all domains using a date-based ranking method positively influences the BiLSTM and RNN model performance - with \textit{evidence-based date ranking} (+2.73/+3.06\% Micro F1) leading to slightly higher results than \textit{claim-based date ranking} (+1.47/+2.18\% Micro F1). We observe similar performance differences between the date-based ranking methods in the DistilBERT model, however, both methods fall behind the base model (-1.34/-7.67\% Micro F1; \textit{evidence-based}/\textit{claim-based date ranking}). For this model, only \textit{evidence-centered distance ranking} yields increased test results (+2.01\% Micro F1). Although this ranking method returns higher test results for the RNN model as well, it is the only temporal ranking method that decreases the BiLSTM model performance (-7.65\% Micro F1). In contrast to the other three temporal ranking methods, \textit{claim-centered distance ranking} does not lead to a substantial performance gain in any of the three fact-checking models (+0.11/-0.72/-3.01\% Micro F1; BiLSTM/RNN/DistilBERT). While performance gains are often limited when applying a single temporal ranking method to all domains, higher results can be obtained by selecting the best performing temporal ranking method per domain based on the development set. As a result, \textit{time-aware evidence ranking} increases model performance by 7.44\% for the BiLSTM model, 8.46\% for the RNN model and 3.63\% for the DistilBERT model (Micro F1).


%% file: discussion.tex
\section{Discussion}

Introducing time-awareness in a fact-checking model by directly optimizing evidence ranking using temporal ranking methods positively influences classification performance. Moreover, time-aware evidence ranking consistently outperforms search engine evidence ranking. This suggests that the temporal ranking methods themselves - and not merely the act of direct evidence ranking optimization - lead to higher results. 
\subsection{Do Time-Aware Models Learn Different Rankings?}

One could ask to which extent the temporal ranking methods actually change the evidence ranking order. To quantify the difference in returned evidence rankings between the base model and the model optimized on one of the temporal ranking methods, we compute the Spearman's rank correlation coefficient $r_s$. We find that the temporal ranking methods consistently change ranking orders in the BiLSTM model as the time-aware evidence rankings are rather weakly correlated with the base rankings ($r_s=.24/.18/.22/.17$; for \textit{evidence-based date ranking}, \textit{claim-based date ranking}, \textit{claim-centered distance ranking} and \textit{evidence-centered distance ranking}, respectively). Although the changes in ranking order are more drastic for the BiLSTM model, the temporal ranking methods have a weaker but still considerable impact on evidence ranking in the RNN model ($r_s = .56/.58/.61/.54$). The lower impact could be attributed to the ranking module's depth: the BiLSTM model’s ranking module consists of twice as many nonlinear fully-connected layers than the RNN’s ranking module. As a result, the deeper ranking module is able to learn more detailed and abstract representations.

While the correlations between time-aware and base ranking order are mainly positive in the BiLSTM and RNN model, evidence rankings are either negatively or completely uncorrelated in the DistilBERT model ($r_s = \minus .23/\minus.35/\minus.02/\minus.07$). In that model, the sentence encoder yields both a label probability distribution and a joint representation for a given claim and evidence snippet. For the other two models, the sentence encoder outputs separate hidden representations for claims and evidence snippets, which are later fused to joint representations. The label scoring module then infers label scores for each joint representation. It can thus be argued that the joint representations in the DistilBERT model are more label-aware, which in turn lead to a more label-biased ranking module that possibly runs counter to a time-aware one. Hence the negative correlations between time-aware and base rankings. We can conclude that the temporal ranking methods indeed cause changes in evidence ranking order, although their impact varies with each model. 
\subsection{What Sparks Time-Aware Ranking Success in a Domain?}

The temporal ranking methods affect overall model performance to various extents: the BiLSTM and RNN model mainly prefer date-based ranking methods, while the DistilBERT model's performance only increases using the evidence-centered distance ranking. Moreover, there is not a single temporal ranking method that increases veracity prediction performance across all models. Similar observations can be made on a by-domain level.
Some domains benefit from all four temporal ranking methods, while others consistently perform worse or are not even affected at all. 

Whether or not time-aware evidence ranking affects a domain's model performance might be ascribed to the share of evidence snippets with retrievable timestamps in that domain. An evidence snippet's computed ranking score can only be optimized by the learning-to-rank loss function if its timestamp can be extracted and, in case of \textit{claim-based date ranking}\footnote{When ranking is constrained using the \textit{claim-based date ranking} method, only evidence snippets with \({\Delta}t \leq 0\) are ranked in the simulated ground-truth rankings.}, is included in the method-specific ground-truth evidence ranking. A large share of time-grounded evidence snippets enables the model to learn the expected rankings in a more direct and constrained manner, while a small share allows more flexible ranking learning. Consequently, it could be argued that the temporal ranking methods influence domains with a large share of time-grounded evidence snippets (\textit{farg}, \textit{chct}, \textit{wast}, \textit{goop}) more strongly and positively than those with a small share (\textit{clck}, \textit{faly}, \textit{fani}). However, that argument is refuted as several small-share domains have their performance increased by large margins, and large-share domains do not consistently benefit from time-aware evidence ranking to a great extent.
These findings suggest that the effectiveness of our temporal ranking methods does not rely on a large amount of time-grounded evidence. 

Another possible cause of the inter-domain differences in time-aware ranking effect is the time-sensitivity of those domains. We hypothesize that domains tackling claims on time-sensitive subjects benefit more from time-aware evidence ranking than those discussing time-insensitive claims. We retrieve the categories\footnote{Categories are given as claim metadata.} from several domains and analyze their time-sensitivity. The analysis confirms our hypothesis: domains which mainly tackle time-sensitive subjects such as politics, economy, climate and entertainment (\textit{abbc}, \textit{para}, \textit{thet}) benefit more from time-aware evidence ranking than domains discussing both time-sensitive and time-insensitive subjects such as food, language, humor and animals (\textit{snes}, \textit{tron}). We can therefore conclude that relating evidence relevance with time and ranking evidence snippets accordingly is beneficial for time-sensitive claims.
\subsection{Do Certain Evidence Distributions Prefer Specific Ranking Methods?}

We observe not only inter-model and inter-domain but also inter-method differences.
Regarding date-based ranking, a domain or model preference for either evidence-based or claim-based date ranking might depend on the share of evidence posted after the claim date. 
If an evidence set mainly consists of later-posted evidence, the ranking of only a few evidence snippets is directly optimized with the claim-based date ranking method, leaving the model to indirectly learn the ranking scores of the others. In that case, the evidence-based method might be favored over the claim-based method. However, the share of later-posted evidence is not consistently different in domains preferring evidence-based date ranking than in domains favoring claim-based date ranking.

Concerning distance-based ranking, evidence and claim (claim-centered distance ranking), and evidence and evidence (evidence-centered distance ranking) are more likely to discuss the same topic when they are published around the same time. The distance-based ranking methods would thus increase classification performance for domains in which the dispersion of evidence snippets in the claim-specific evidence sets is small. We measure the temporal dispersion of each evidence set using the standard deviation in domains which mainly favor distance-based ranking over date-based ranking (\textit{afck}, \textit{faan}, \textit{pomt}, \textit{thal}; Group 1), and vice versa (\textit{chct}, \textit{vogo}; Group 2). 
We then check whether the domains in these groups display similar dispersion values. Kruskal-Wallis H tests indicate that dispersion values statistically differ between domains in the same group (Group 1: $H = 258.63, p < 0.01$; Group 2: $H = 192.71, p < 0.01$). Moreover, Mann-Whitney U tests on domain pairs suggest that inter-group differences are not consistently larger than intra-group differences (e.g., \textit{thal-chct}: $Mdn = 337.53, Mdn = 239.26, p = 0.021 > 0.01$;  \textit{thal-pomt}: $Mdn = 337.53, Mdn = 661.25, p = 9.95e^{-5} < 0.01$). Therefore, the hypothesis that small evidence dispersion causes a preference for distance-based ranking methods is rejected. 

\subsection{On Claim-Specific Ranking and Temporal Semantics}

In the experiments, the ground-truth rankings of all evidence sets within a domain are simulated using a single temporal ranking method. Our approach successfully improves classification performance - especially when selecting the best temporal ranking method per domain (\textit{time-aware evidence ranking}) - and significantly influences ranking order. Nevertheless, it could be argued that additional performance increases could be gained from inferring the most beneficial temporal ranking method for each claim individually. This would require a more in-depth reasoning over the temporal semantics and importance hierarchy of information contained within a claim. 
Consider the following claims taken from the \textit{abbc} domain:
\begin{enumerate}
    \item \textit{Australians now need to visit a doctor to obtain codeine after the drug was taken off pharmacy shelves and made available only with a prescription.} (claim date: 23 March 2018)
    \item \textit{``Indigenous children at the moment are 10 times more likely to be living out of home right now," Greens Senator Sarah Hanson-Young said on the ABC's Q\&A.} (claim date: 13 April 2016)
    \item \textit{On September 4, 2013, the Liberal Party promised that, if elected, ``a Coalition government will establish a new `seniors employment incentive payment' of \$3,250 for employers that hire mature workers aged 50 or older and keep them on for at least six months."} (claim date: 27 July 2014)
\end{enumerate}
In the first claim, temporal cues \textit{now} in the main clause (\textit{Australians now need to visit a doctor to obtain codeine}) and \textit{after} in the subordinate clause (\textit{after the drug was taken off pharmacy shelves and made available only with a prescription}) indicate that the situation described in the main clause (a) is happening at claim time and (b) occurred after the two events described in the subordinate clause. Moreover, the temporal cues mark a causal discourse relation where the main clause is caused by the subordinate clause. This causal relation can be considered world knowledge as prescribed drugs can only be legally obtained after a doctor's visit. As a result, a fact-checking model can rely on either one of the clauses to predict the veracity of the entire claim. In terms of the temporal relevance of the claim's evidence set, evidence published either around the same time as the claim or after the claim are more relevant to predict the veracity of the claim. As \textit{now} in the main clause implies that the Australian policy for codeine was different before the claim date, old evidence would provide incorrect information - even if the evidence is provided by reliable sources. That issue specially highlights the importance of temporal semantics in fact-checking.

In contrast to the first claim, there is a greater importance hierarchy between the clauses of the second and third claim. On the one hand, a fact-checking model can decide to focus on the entire claim and verify whether \textit{Greens Senator Sarah Hanson-Young} and \textit{the Liberal Party} actually verbalized the respective subordinating clauses. For the second claim, the model needs to establish when that specific ABC's Q\&A session with the senator occurred and look for evidence that reports on the event, preferable published on the same day or the day after the event. Given the past tense of the main verb (\textit{said}), the event arguably occurred before the claim data, so the model should look for older evidence. However, the definite article \textit{the} in \textit{the ABC's Q\&A} suggests that the main clause refers to the latest Q\&A session. So the model should not look too far in the past. For the third claim, an explicit date is provided (\textit{September 4, 2013}). A model can thus look for evidence published on that specific day to check whether the \textit{Liberal Party} made that promise.

Instead of focusing on the entire claim, a fact-checking model could also choose to verify the information given in the subordinating clauses. This would require an even more thorough temporal reasoning as the model should not only place the event or situation in the subordinating clause in time, but also establish the temporal semantics of the main clause and the temporal relation between the two clauses. The temporal cues \textit{at the moment} and \textit{right now} in the subordinating clause of the second claim (\textit{``Indigenous children at the moment are 10 times more likely to be living out of home right now,"}) suggest that the situation is happening at claim time. However, the past tense of the verb (\textit{said}) in the main clause puts the entire subordinating clause in the past. As a result, evidence from the time of the Q\&A session will be considered more relevant than evidence published around the claim time. For the third claim, the subordinating clause refers to an uncertain event that might happen in the future given the future tense \textit{will establish} and the phrase \textit{if elected}. It requires extensive world knowledge to check the information in the subordinating clause. Firstly, the model needs to check whether the \textit{Liberal Party} was indeed elected and a \textit{Coalition government} was formed. It then needs to establish whether or not that government agreed on the \textit{new seniors employment incentive payment}. Lastly, the model can take it a step further and take into account evidence published after the claim date to check whether that government was able to keep its promise before the end of its administration. As the subordinating clause might not be true at claim time, its veracity could have changed over time. 

Future work can zoom in on the temporal semantics at claim level and apply various temporal reasoning methodologies to extract temporal information and construct event timelines from text \cite{leeuwenberg2018temporal}. A comprehensive overview of temporal reasoning methodologies can be found in \citet{leeuwenberg2019survey}. Based on a claim's temporal semantics, an appropriate temporal ranking method can be selected to simulate a claim-specific ground-truth evidence ranking.

\subsection{Limitations}

The impact and success of the temporal ranking methods still depend on the informativeness of the given Web documents which serve as evidence to the claim. As the Web documents are automatically crawled from the Internet given the claim as query to the Google Search API, it is not guaranteed that they are all useful for predicting that claim's veracity. 
If a large number of Web documents in the evidence set do not contain useful information for refuting or supporting the claim, enforcing a temporal ranking will then have little to no effect on model performance. To analyse this, we randomly pick a claim with a large number of evidence snippets with a retrievable timestamp, but for which both base and time-aware models consistently return an incorrect veracity label (Table \ref{tab:close_look}). The veracity and semantics of the claim are time-dependent (i.e., trade balance varies over time), but the majority of Web documents are irrelevant to the claim. Consequently, a fact-checking model cannot make an informed veracity prediction - independent of how it ranks the evidence. 

It should generally be noted that the Google Search engine has been evaluated as being highly effective in returning diverse results for queries \cite{journals/ires/WuZX19}. In future work, to obtain a larger set of evidence snippets for temporal ranking, one could experiment with retrieving more evidence snippets per query, use multiple search engines or automatically formulate multiple search queries per claim, as opposed to merely submitting the claim as a query.

\begin{table*}[]
    \fontsize{7.5}{8.5}\selectfont
    \centering
    \begin{tabular}{p{2cm}p{10cm}}
        \toprule
        Timestamp & \textbf{Claim} \\
        \toprule
        Mar 20, 2018 & `Claim on Scotland’s positive trade balance compared to UK nations' \\
        \toprule
        & \textbf{Evidence Set} \\
        \toprule
        Mar 20, 2018 & Mar 20, 2018 ... Claim on Scotland's positive trade balance compared to UK nations is ... with a  new Mostly False verdict and correction note, and re-posted to ... \\
        \midrule
        / & Advice, and information so that anyone can check the claims we hear  about ... on Scotland's positive trade balance compared to UK nations is Mostly  False \\
        \midrule
        Mar 10, 2018 & Mar 10, 2018 ... Claim: The US is suffering from a trade imbalance, with a trade ... Reality Check  verdict: The President is incorrect about the US trade deficit - it was \$566bn (£ 410bn) in 2017. ... declined in most major western industrialised nations over the  past ... 9 Two British soldiers injured in Islamic State attack in Syria ... \\
        \midrule
        Sep 7, 2016 & Sep 7, 2016 ... Adam Smith's 1776 classic "Wealth of Nations" may have had the ... What was the  most important document published in 1776? ... Smith, a Scottish philosopher by  trade, wrote the book to upend the mercantilist system. ... fallacies in an argument  that is framed as the invisible hand versus the government. \\
        \midrule
        Apr 17, 2018 & Apr 17, 2018 ... Everything You've Been Told About Government Debt Is Wrong ... usually  presume that the government will run a primary surplus (the excess of ... For  example, during the era of relative peace following 1960, decadal ... Using this as  a parameter, Barrett estimates a safe debt/GDP level for the U.K. of 140\%. \\
        \midrule
        Apr 20, 2018 & Apr 20, 2018 ... President Donald Trump's trade policy leaves international ... trade system, which  I'd argue has benefited the nation economically overall ... of thought most  economists believe Adam Smith extinguished after he ... Second, imposing new  and higher tariffs on imports won't make the U.S. trade deficit go away. \\
        \midrule
        Jun 29, 2016 & Jun 29, 2016 ... Similar information for devolved administrations are available at Scotland: Fire  and Rescue Statistics ... FIRE0103: Fires attended by fire and rescue services by  nation and ... FIRE0104: Fire false alarms by reason for false alarm, England (MS  .... FIRE1303: Firefighters' pension surplus and deficit (MS Excel... \\
        \midrule
        Aug 6, 2017 & Aug 6, 2017 ... After the bitter referendums over Scottish independence and ... Judged on hard  metrics, confidence in UK media has fallen .... Remainers detected dangerous  instances of false balance, most notoriously when a poll found that 88\% of UK ...  who said that Brexit would not damage trade and the UK economy. \\
        \midrule
        / & ... the failure of people in the rich nations  to make any significant sacrifices in ... It is not simply the absence of charity, let  alone of moral saintliness: It is wrong, and one cannot claim to be a morally  decent ... without thereby sacrificing anything of comparable moral importance,  we... \\
        \bottomrule
    \end{tabular}
    \caption{Claim and accompanying evidence set from the MultiFC dataset \cite{augenstein2019multifc} for which both base and time-aware models consistently predict an incorrect veracity label.}
    \label{tab:close_look}
\end{table*}



%% file: conclusion.tex
\section{Conclusion}

Introducing time-awareness in evidence ranking arguably leads to more accurate veracity predictions in fact-checking models -- especially when they deal with claims about time-sensitive subjects such as politics and entertainment. These performance gains also indicate that evidence relevance should be approached more diversely instead of merely associating it with the semantic similarity between claim and evidence. By integrating temporal ranking constraints in neural architectures via appropriate loss functions, we show that fact-checking models are able to learn time-aware evidence rankings in an elegant, yet effective manner. To our knowledge, evidence ranking optimization using a dedicated ranking loss has not been studied before in the context of fact-checking. Whereas this study is limited to integrating time-awareness in the evidence ranking as part of automated fact-checking, future research could build on these findings to explore the impact of time-awareness at other stages of fact-checking, e.g., document retrieval or evidence selection, and in domains beyond fact-checking. Alternatively, the analogy with spatial relevance can be explored by adopting similar spatial ranking methods for space-aware evidence ranking.

%% file: supplementary_material.tex
\begin{table}[h!]
    \centering
    \small
    \begin{tabular}{c|c}
        \toprule
        \multicolumn{2}{c}{Hyperparameter settings} \\
        \midrule
        batch size &  32\\
        epochs & 150 \\
        word embedding size & 300 \\
        word embedding weight initialization & Xavier uniform \\
        BiLSTM \# layers & 2 \\
        BiLSTM skip-connections & concatenation \\
        BiLSTM hidden size & 128 \\
        CNN out channels & 32 \\
        CNN kernel size & 32 \\
        CNN activation function & ReLU \\
        ranking FC weight initialization & Xavier uniform \\
        ranking FC (1) activation function & ReLU \\
        ranking FC (2) activation function & Leaky ReLU \\
        label embedding size & 16 \\
        label embedding weight initialization & Xavier uniform \\
        label embedding FC weight initialization & Xavier uniform \\
        label embedding FC activation function & Leaky ReLU \\
        RMSProp learning rate & 0.0002 \\
        Adam learning rate & 0.001 \\
        \bottomrule
    \end{tabular}
    \caption{Hyperparameter settings \textbf{BiLSTM model}.}
    \label{tab:hyper_multifc}
\end{table}

\begin{table}[ht!]
    \centering
    \small
    \begin{tabular}{c|c}
        \toprule
        \multicolumn{2}{c}{Hyperparameter settings} \\
        \midrule
        batch size &  32\\
        epochs & 150 \\
        word embedding size & 300 \\
        word embedding weight initialization & Xavier uniform \\
        RNN \# layers & 2 \\
        RNN dropout & 0.1 \\
        RNN hidden size & 128 \\
        CNN out channels & 32 \\
        CNN kernel size & 32 \\
        CNN activation function & ReLU \\
        ranking FC weight initialization & Xavier uniform \\
        ranking FC activation function & Sigmoid \\
        label FC weight initialization & Xavier uniform \\
        label FC activation function & Leaky ReLU \\
        RMSProp learning rate & 0.0002 \\
        Adam learning rate & 0.001 \\
        \bottomrule
    \end{tabular}
    \caption{Hyperparameter settings \textbf{RNN model}.}
    \label{tab:hyper_rnn}
\end{table}

\begin{table}[ht!]
    \centering
    \small
    \begin{tabular}{c|c}
        \toprule
        \multicolumn{2}{c}{Hyperparameter settings} \\
        \midrule
        batch size &  32\\
        epochs & 10 \\
        DistilBERT configuration & distilbert-base-uncased \\
        DistilBERT model & DistilBERT for \\
         & sequence classification \\
        ranking FC weight initialization & Xavier uniform \\
        ranking FC (1) activation function & ReLU \\
        ranking FC (2) activation function & Leaky ReLU \\
        AdamW learning rate & 0.0005 \\
        Adam learning rate & 0.01 \\
        \bottomrule
    \end{tabular}
    \caption{Hyperparameter settings \textbf{DistilBERT model}.}
    \label{tab:hyper_distilbert}
\end{table}

%% file: supplementary_material_B.tex
\begin{table*}[!htpb]
    \fontsize{7.5}{8.5}\selectfont
    \setlength{\tabcolsep}{1.5pt}
    \centering
    \begin{tabular}{ccc||cc||cccccccc||cc}
        \toprule  & 
        \multicolumn{2}{c}{\textbf{Base}} & \multicolumn{2}{c}{\textbf{Search}} & \multicolumn{2}{c}{\textbf{Evidence}} & \multicolumn{2}{c}{\textbf{Claim}} &
        \multicolumn{2}{c}{\textbf{Claim}} &
        \multicolumn{2}{c}{\textbf{Evidence}} &
        \multicolumn{2}{c}{\textbf{Time-Aware}}\\
        & \multicolumn{2}{c}{\textbf{Model}} & \multicolumn{2}{c}{\textbf{Ranking}} & \multicolumn{2}{c}{\textbf{Date}} & \multicolumn{2}{c}{\textbf{Date}} &
        \multicolumn{2}{c}{\textbf{Distance}} &
        \multicolumn{2}{c}{\textbf{Distance}} &
        \multicolumn{2}{c}{\textbf{Ranking}}\\
        \cmidrule{2-15}
         & Micro & Macro & Micro & Macro & Micro & Macro & Micro & Macro & Micro & Macro & Micro & Macro & Micro & Macro \\
         \midrule
        abbc & .4815 & .3111 & .2963 & .2301 &  .5000 & .2278 & .5556 & .4233 & .4444 & .2133 & .4444 & .3139 & \underline{.5556} & \underline{.4233} \\
        afck & .2500 & .1191 & .2778 & .0725 &  .2222 & .1215 & .2500 & .1252 & .3056 & .1546 & .3056 & .1891  & \underline{.3056} & \underline{.1891} \\
        bove & 1. & 1.  & 1. & 1. & 1. & 1. & 1. & 1. & 1. & 1. & 1. & 1. & 1. & 1. \\
        chct & .4000 & .2337 & .4000 & .1429 & .4000 & .2122 & .4250 & .2347 & .6250 & .3365 & .4500 & .2836 & \underline{.6250} & \underline{.3365} \\
        clck & .1667 & .1333  & .1667 & .0952 & .6667 & .4545 & .8333 & .6222 & .5000 & .3590 & .5000 & .2500 & \underline{.8333} & \underline{.6222} \\
        faan & .5000 & .3529  & .4545 & .2083 & .4091 & .2000 & .3636 & .3228 & .5000 & .4442 & .4545 & .3472 & .5000 & \underline{.4442} \\
        faly & .5714 & .1818  & .5000 & .1750 & .5714 & .1905 & .6429 & .1957 & .5000 & .2321 & .5000 & .2321 & \underline{.6429} & \underline{.1957} \\
        fani & 1. & 1.  & 1. & 1. & 1. & 1. & 1. & 1. & 1. & 1. & 1. & 1. & 1. & 1. \\
        farg & .6923 & .1023  & .6923 & .1023 & .7115 & .1803 & .6923 & .1023 & .6923 & .1023 & .6538 & .0988 & \underline{.7115} & \underline{.1803} \\
        goop & .8344 & .1516  & .8344 & .1516 & .8344 & .1516 & .8344 & .1516 & .8344 & .1516 & .8344 & .1516 & .8344 & .1516 \\
        hoer & .3657 & .1120  & .3731 & .0446 & .3806 & .1260 & .3731 & .1152 & .3657 & .1292 & .3433 & .0982 & \underline{.3806} & \underline{.1260} \\
        huca & .3333 & .2222  & .1667 & .2222 & .3333 & .1667 & .1667 & .0952 & .1667 & .0952 & .5000 & .2500 & \underline{.5000} & \underline{.2500} \\
        mpws & .5000 & .2222  & .1250 & .0833 & .6250 & .4148 & .3750 & .3241 & .3750 & .1818 & .6250 & .4809 & \underline{.6250} & \underline{.4809} \\
        obry & .3571 & .1471  & .4286 & .2755 & .5000 & .2847 & .4286 & .3056 & .0714 & .0357 & .2857 & .1641 & \underline{.5000} & \underline{.2847} \\
        para & .2188 & .1033  & .1875 & .1104 & .1875 & .1061 & .2813 & .1415 & .2188 & .1643 & .3438 & .2993 & \underline{.3438} & \underline{.2993} \\
        peck & .5000 & .3397  & .5833 & .2456 & .5833 & .3679 & .7500 & .5095 & .5000 & .2222 & .5833 & .3679 & \underline{.7500} & \underline{.5095} \\
        pomt & .1955 & .0792  & .1750 & .0753 & .2196 & .1578 & .2286 & .1651 & .2152 & .1627 & .2402 & .1699 & \underline{.2402} & \underline{.1699} \\
        pose & .4110 & .1335  & .4178 & .0982 & .4178 & .0982 & .4041 & .1908 & .3904 & .1279 & .4178 & .1864 & \underline{.4178} & \underline{.1864} \\
        ranz & 1. & 1.  & 1. & 1. & 1. & 1. & 1. & 1. & 1. & 1. & .5000 & .3333 & 1. & 1. \\
        snes & .5646 & .0601  & .5646 & .0601 & .5646 & .0601 & .5646 & .0601 & .5646 & .0601 & .5554 & .0796 & .5646 & .0601 \\
        thal & .5000 & .2688  & .5556 & .1786 & .5000 & .1667 & .5556 & .1786 & .3333 & .1691 & .5556 & .1786 & \underline{.5556} & .1786 \\
        thet & .5625 & .2419  & .4375 & .1167 & .5625 & .1636 & .2500 & .0784 & .5000 & .1333 & .5625 & .1500 & .5625 & .1636 \\
        tron & .4767 & .0258  & .4767 & .0258 & .4767 & .0398 & .4767 & .0258 & .4651 & .0326 & .4680 & .0360 & .4767 & \underline{.0398} \\
        vees & .6613 & .2381  & .7097 & .2075 & .5806 & .2328 & .5806 & .1875 & .7097 & .2075  & .7097 & .2115 & \underline{.7097} & \underline{.2115} \\
        vogo & .3750 & .0779  & .3906 & .1735 & .4375 & .2467 & .4531 & .1979 & .4531 & .2286 & .4063 & .1769 & \underline{.4531} & \underline{.2286} \\
        wast & .2188 & .1387  & .2188 & .0718 & .2500 & .2467 & .2188 & .2595 & .2188 & .0757 & .1875 & .1343 & \underline{.2500} & \underline{.2467} \\
        \midrule
        avg. & .5053 & .2691  & .4782 & .2372 & \underline{.5359} & \underline{.2905} & \underline{.5271} & \underline{.3082} & .4981 & \underline{.2700}  & \underline{5356} & \underline{.3019} & \underline{.5899} & \underline{.3453} \\
        \bottomrule
    \end{tabular}
    \caption{Overview of classification test results (Micro F1 and Macro F1) for \textbf{RNN model - PT + FT}, with improvements over base model results \underline{underlined}. The last column contains the highest results returned by the four temporal ranking methods.}
    \label{tab:results_rnn}
\end{table*}

\begin{table*}[!htpb]
    \fontsize{7.5}{8.5}\selectfont
    \setlength{\tabcolsep}{1.5pt}
    \centering
    \begin{tabular}{ccc||cc||cccccccc||cc}
        \toprule  & 
        \multicolumn{2}{c}{\textbf{Base}} & \multicolumn{2}{c}{\textbf{Search}} & \multicolumn{2}{c}{\textbf{Evidence}} & \multicolumn{2}{c}{\textbf{Claim}} &
        \multicolumn{2}{c}{\textbf{Claim}} &
        \multicolumn{2}{c}{\textbf{Evidence}} &
        \multicolumn{2}{c}{\textbf{Time-Aware}}\\
        & \multicolumn{2}{c}{\textbf{Model}} & \multicolumn{2}{c}{\textbf{Ranking}} & \multicolumn{2}{c}{\textbf{Date}} & \multicolumn{2}{c}{\textbf{Date}} &
        \multicolumn{2}{c}{\textbf{Distance}} &
        \multicolumn{2}{c}{\textbf{Distance}} &
        \multicolumn{2}{c}{\textbf{Ranking}}\\
        \cmidrule{2-15}
         & Micro & Macro & Micro & Macro & Micro & Macro & Micro & Macro & Micro & Macro & Micro & Macro & Micro & Macro \\
         \midrule
        abbc & .5370 & .4181 & .5370 & .2329 &  .5370 & .2329 & .5370 & .2329 & .5370 & .2329 & .5370 & .2329 & .5370 & .2329 \\
        afck & .1667 & .0987 & .1994 & .0903 &  .1944 & .0903 & .1944 & .0903 & .1389 & .0663 & .2778 & .1273  & \underline{.2778} & \underline{.1273} \\
        bove & 1. & 1.  & 1. & 1. & 1. & 1. & 1. & 1. & 1. & 1. & 1. & 1. & 1. & 1. \\
        chct & .5250 & .2854 & .4000 & .1429 & .4500 & .1552 & .4250 & .2121 & .4000 & .2321 & .5250 & .2846 & .5250 & .2846 \\
        clck & .1667 & .0952  & .1667 & .0952 & .1667 & .0952 & .1667 & .0952 & .3333 & .1667 & .5000 & .2222 & \underline{.5000} & \underline{.2222} \\
        faan & .4091 & .1667  & .4545 & .2083 & .4545 & .2083 & .1818 & .1159 & .7273 & .5593 & .7273 & .5388 & \underline{.7273} & \underline{.5593} \\
        faly & .8571 & .4762  & .0714 & .0333 & .8571 & .4521 & .2143 & .0882 & .8571 & .4521 & .8751 & .4521 & .8571 & .4521 \\
        fani & 1. & 1.  & 1. & 1. & 1. & 1. & 1. & 1. & 1. & 1. & 1. & 1. & 1. & 1. \\
        farg & .6923 & .1023  & .6923 & .1023 & .6923 & .1023 & .6923 & .1023 & .6923 & .1023 & .6538 & .0988 & .6923 & .1023 \\
        goop & .8344 & .1516  & .8344 & .1516 & .8344 & .1516 & .8344 & .1516 & .8344 & .1516 & .8344 & .1516 & .8344 & .1516 \\
        hoer & .5075 & .1863  & .3731 & .0776 & .3731 & .0776 & .3731 & .0776 & .3731 & .0776 & .4851 & .1671 & .4851 & .1671 \\
        huca & .5000 & .2857  & .1667 & .0952 & .5000 & .2222 & .3333 & .1667 & .1667 & .0952 & .1667 & .0952 & .5000 & .2222 \\
        mpws & .5000 & .2222  & .5000 & .2222 & .5000 & .2222 & .5000 & .2222 & .5000 & .2222 & .6250 & .4578 & \underline{.6250} & \underline{.4578} \\
        obry & .4286 & .1500  & .4286 & .1500 & .4286 & .1500 & .4286 & .2483 & .4286 & .1500 & .4286 & .1500 & .4286 & \underline{.2483} \\
        para & .2500 & .1435  & .1875 & .0526 & .2500 & .0667 & .2813 & .1530 & .2500 & .0667 & .1250 & .0370 & \underline{.2813} & \underline{.1530} \\
        peck & .6667 & .4652  & .5833 & .2456 & .8333 & .5879 & .5833 & .2456 & .5833 & .2456 & .5833 & .2456 & \underline{.8333} & \underline{.5879} \\
        pomt & .2482 & .1263  & .1938 & .0361 & .2464 & .1635 & .1732 & .0328 & .2536 & .1634 & .2545 & .1650 & \underline{.2545} & \underline{.1650} \\
        pose & .4178 & .0982  & .4178 & .0982 & .4178 & .0982 & .4178 & .0982 & .4178 & .0982 & .4178 & .0982 & .4178 & .0982 \\
        ranz & 1. & 1.  & 1. & 1. & 1. & 1. & 1. & 1. & 1. & 1. & 1. & 1. & 1. & 1. \\
        snes & .5646 & .0601  & .5646 & .0601 & .5646 & .0601 & .5646 & .0601 & .5646 & .0601 & .5645 & .0601 & .5646 & .0601 \\
        thal & .5556 & .0601  & .5556 & .1786 & .3333 & .1043 & .5556 & .1786 & .1111 & .0500 & .5000 & .2273 & .5556 & \underline{.1786} \\
        thet & .5625 & .1440  & .5625 & .1440 & .5625 & .1440 & .5625 & .1440 & .5625 & .1500 & .5625 & .1440 & .5625 & .1440 \\
        tron & .4767 & .0258  & .4767 & .0258 & .4767 & .0258 & .4767 & .0258 & .4767 & .0258 & .4767 & .0258 & .4767 & .0258 \\
        vees & .7097 & .2075  & .7097 & .2075 & .7097 & .2115 & .7097 & .2075 & .7097 & .2075  & .7097 & .2075 & .7097 & \underline{.2115} \\
        vogo & .5313 & .2199  & .1719 & .1567 & .5000 & .2146 & .1250 & .1676 & .5313 & .2199 & .5156 & .2098 & .5313 & .2199 \\
        wast & .3438 & .2546  & .2188 & .0800 & .2188 & .0800 & .1250 & .1008 & .2188 & .0800 & .2188 & .0800 & .2188 & .0800 \\
        \midrule
        avg. & .5558 & .2909  & .4782 & .2372 & \underline{.5359} & \underline{.2905} & \underline{.5271} & \underline{.3082} & .4981 & \underline{.2700}  & \underline{5356} & \underline{.3019} & \underline{.5921} & \underline{.3135} \\
        \bottomrule
    \end{tabular}
    \caption{Overview of classification test results (Micro F1 and Macro F1) for \textbf{DistilBERT model - PT + FT}, with improvements over base model results \underline{underlined}. The last column contains the highest results returned by the four temporal ranking methods.}
    \label{tab:results_distilbert}
\end{table*}

\begin{table*}[!htb]
    \fontsize{7.5}{8.5}\selectfont
    \setlength{\tabcolsep}{1.5pt}
    \centering
    \begin{tabular}{ccc||cccccccc}
        \toprule  & 
        \multicolumn{2}{c}{\textbf{}} & \multicolumn{2}{c}{\textbf{All Temporal}} & \multicolumn{2}{c}{\textbf{Evidence Date}} & \multicolumn{2}{c}{\textbf{Claim Distance}} &
        \multicolumn{2}{c}{\textbf{Best Temporal}}\\
        & \multicolumn{2}{c}{\textbf{Base Model}} & \multicolumn{2}{c}{\textbf{Ranking Methods}} & \multicolumn{2}{c}{\textbf{+ Claim Date}} & \multicolumn{2}{c}{\textbf{+ Evidence Distance}} &
        \multicolumn{2}{c}{\textbf{Ranking Method}}\\
        
        \cmidrule{2-11}
         & Micro & Macro & Micro & Macro & Micro & Macro & Micro & Macro & Micro & Macro \\
         \midrule
        abbc & .3148 & .2782 & \underline{.5556} & \underline{.4083} & \underline{.5185} & .2276 & \underline{.5185} & .2276 & \underline{.5185} & .2276 \\
        afck & .2222 & .1385 & \underline{.2500} & .1145 & \underline{.2778} & .1280  & .2222 & .1009 &  \underline{.3333} & \underline{.1517} \\
        bove & 1. & 1. & 1. &  1. & 1. & 1.  & 1. & 1.  & 1. & 1. \\
        chct & .5500 & .2967 & .5250 & .2877 & .4250 & .2297  & .4250 & .2310 & .4750 & .2613 \\
        clck & .5000 & .2500 & .5000 & .2500 & .5000 & .2500 & .5000 & .2500 & .5000 & .2500 \\
        faan & .5000 & .5532 & \underline{.5455} & .1609 & .4091 & .2799 & .3182 & .1728 & .4091 & .2918 \\
        faly & .6429 & .2045 & .5000 & \underline{.2321} & .6429 & .6429 & .6429 & .2045 & .6429 & .2045 \\
        fani & 1. & 1.& 1.& 1. & 1. & 1. & 1. & 1. & 1. & 1. \\
        farg & .6923 & .1023 & .6923 & .1023 & .6923 & .1023 & .6923 & .1023 & .6923 & .1023  \\
        goop & .8344 & .1516 & .8344 & .1516 & .8311 & .1516 & .8013 & \underline{.1583} & .8344 & .1516 \\
        hoer & .4104 & .2091 & \underline{.4179} & \underline{.2226} & .3955 & .1742 & .4104 & .1926 & \underline{.4552} & \underline{.2453} \\
        huca & .5000 & .2857 & .5000 & .2857 & .5000 & .2857 & .5000 & .2857 & \underline{.6667} & \underline{.5333} \\
        mpws & .8750 & .6316 & .8750 & .6316 & .8750 & .6316  & .8750 & .6316 & .8750 & .6316 \\
        obry & .5714 & .3520 & .3571 & .1316 & .5714 & \underline{.3716} & .2143 & .0882 & .5714 & \underline{.3716} \\
        para & .1875 & .1327 & \underline{.3750} & \underline{.1692} & \underline{.3750} & \underline{.1692} & \underline{.3750}  & \underline{.1692} & \underline{.4375} & \underline{.2010} \\
        peck & .5833 & .2593 & \underline{.6667} & \underline{.4082} & \underline{.6667} & \underline{.4082} & \underline{.6667} & \underline{.4082} & \underline{.9167} & \underline{.6316} \\
        pomt & .2143 & .2089 & .1777 & .0579 & .1804 & .0597 & .1804 & .0593 & .1964 & .0637 \\
        pose & .4178 & .0987 & .4178 & \underline{.1002} & .4178 & .0987  & .4041 & .0964 & \underline{.4247} & \underline{.2080} \\
        ranz & 1. & 1. & 1. & 1. & 1. & 1. & 1. & 1. & 1. & 1. \\
        snes & .5646 & .0601 & .5631 & .0601 & .5646 & \underline{.0602} & .5646 & \underline{.0602} & .5646 & .0601 \\
        thal & .5556 & .2756 & .3333 & .1143 & .2778 & .1250 & .5556 & \underline{.3405} & .5556 & \underline{.3600} \\
        thet & .3750 & .2593 & \underline{.6250} & \underline{.3556}  & \underline{.6250} & \underline{.3556} & \underline{.6250} & \underline{.3007} & \underline{.6250} & \underline{.3556} \\
        tron & .4767 & .0258 & .4767 & .0258 & .4767 & .0258 & .4593 & \underline{.0321} & .4767 & \underline{.0366} \\
        vees & .7097 & .2075 & .7097 & .2075 & .7097 & .2075 & .7097 & .2075 & \underline{.7258} & \underline{.3021} \\
        vogo & .5000 & .2056 & .4375 & .1822 & .4688 & .1883 & \underline{.5156} & \underline{.2491} & \underline{.5469} & \underline{.2463} \\
        wast & .1563 & .0935 & \underline{.2188} & .0757 & \underline{.2188} & .0737 & .0938 & .0400 & \underline{.3438} & \underline{.2785} \\
        \midrule
        avg. & .5521 & .3185 & \underline{.5598} & .2975 & \underline{.5623} & \underline{.3172} & .5488 & .2926 & \underline{.6072} & \underline{.3525} \\
        \bottomrule
    \end{tabular}
    \caption{Overview of classification test results (Micro F1 and Macro F1) for \textbf{BiLSTM model - Only FT}, with improvements over base model results \underline{underlined}. We present three optimization approaches, in addition to the approach used in the main paper. Instead of optimizing the evidence ranking based on single temporal ranking methods, all four temporal ranking methods (evidence-based date ranking, claim-based date ranking, claim-based distance and evidence-based distance; \textbf{All Temporal Ranking Methods}) are considered, and the evidence ranking is optimized based on the sum of the four ranking losses. We also explore a combination of the date-based ranking losses (\textbf{Evidence Date + Claim Date}) and the distance-based ranking losses (\textbf{Claim Distance + Evidence Distance}). The lower results motivate our choice for the optimization approach presented in the main paper (i.e., optimization based on single temporal ranking methods and selecting the best method for each domain; \textbf{Best Temporal Ranking Method}).}
    \label{tab:results_approaches}
\end{table*}